\documentclass{article}

\PassOptionsToPackage{numbers, compress}{natbib}
\usepackage[final]{neurips_2019_emc2_extended}

\usepackage[utf8]{inputenc} % allow utf-8 input
\usepackage[T1]{fontenc}    % use 8-bit T1 fonts
\usepackage{hyperref}       % hyperlinks
\usepackage{url}            % simple URL typesetting
\usepackage{booktabs}       % professional-quality tables
\usepackage{amsfonts}       % blackboard math symbols
\usepackage{nicefrac}       % compact symbols for 1/2, etc.
\usepackage{microtype}      % microtypography

\usepackage{graphicx, subcaption}
\usepackage[justification=centering]{caption}
\usepackage{adjustbox}
\usepackage{scrextend}
\usepackage{algorithm2e}
\usepackage{mathtools}
\usepackage[inline]{enumitem}
\usepackage{todonotes}
\usepackage{amsmath}
\usepackage{ellipsis}
\usepackage{cleveref}

\usepackage{mathtools}

\title{Instant Quantization of Neural Networks using Monte Carlo Methods}

\author{%
  Gon{\c c}alo Mordido\thanks{Equal contribution.}\enspace\thanks{Work done during a research internship at NVIDIA.}
  \\
  Hasso Plattner Institute\\
  Potsdam, Germany\\
  \texttt{goncalo.mordido@hpi.de} \\
  \And
  Matthijs Van Keirsbilck\footnotemark[1]\\
  NVIDIA\\
  Berlin, Germany \\
  \texttt{matthijsv@nvidia.com} \\
  \And
  Alexander Keller \\
  NVIDIA\\
  Berlin, Germany \\
  \texttt{akeller@nvidia.com} \\
}

\begin{document}

\maketitle

\begin{abstract}
    Low bit-width integer weights and activations are very important for efficient inference, especially with respect to lower power consumption.
    We propose to apply Monte Carlo methods and importance sampling to sparsify and quantize pre-trained neural networks without any retraining. We obtain sparse, low bit-width integer representations that approximate the full precision weights and activations. 
    The precision, sparsity, and complexity are easily configurable by the amount of sampling performed.
    Our approach, called Monte Carlo Quantization (MCQ), is linear in both time and space, while the resulting quantized sparse networks show minimal accuracy loss compared to the original full-precision networks. Our method either outperforms or achieves results competitive with methods that do require additional training on a variety of challenging tasks.
\end{abstract}

\section{Introduction}

Developing novel ways of increasing the efficiency of neural networks is of great importance due to their widespread usage in today's variety of applications. Reducing the network's footprint enables local processing on personal devices without the need for cloud services. In addition, such methods allow for reducing power consumption - also in data centers. 
Very compact models can be fully stored and executed on-chip in specialized hardware like for example ASICs or FPGAs.
This reduces latency, increases inference speed, improves privacy concerns, and limits bandwidth cost. 

Quantization methods usually require re-training of the quantized model to achieve competitive results. This leads to an additional cost and complexity. The proposed method, Monte Carlo Quantization (MCQ), aims to avoid retraining by approximating the full-precision weight and activation distributions using importance sampling. The resulting quantized networks achieve close to the full-precision accuracy without any kind of additional training. Importantly, the complexity of the resulting networks is proportional to the number of samples taken.

First, our algorithm normalizes the weights and activations of a given layer to treat them as probability distributions. Then, we randomly sample from the corresponding cumulative distributions and count the number of hits for every weight and activation.
Finally, we quantize the weights and activations by their integer count values, which form a discrete approximation of the original continuous values. 
Since the quality of this approximation relies entirely on (quasi)random sampling, the accuracy of the quantized model is directly dependent on the amount of sampling performed. Thus, accuracy may be traded for higher sparsity and speed by adjusting the number of samples.
On the challenging tasks of image classification, language modeling, speech recognition, and machine translation, our method outperforms or is competitive with existing quantization methods that do require additional training.

\section{Related Work}\label{sec:related_work}
    
    The computational cost of neural networks can be reduced by pruning redundant weights or neurons, which has been shown to work well \cite{deepcompression, mocanu2018scalable, lecun1990optimal}. 
    Alternatively, the precision of the network weights and activations may be lowered, potentially introducing sparsity. 
    Using low precision computations to reduce the cost and sparsity to skip computations allows for efficient hardware implementations~\cite{lin2015neural, dlac}. This is the approach used in this paper.
    
    BinaryConnect ~\cite{courbariaux2015binaryconnect} proposed training with binary weights, while XNOR-Net ~\cite{xnornet} and BNN ~\cite{bnn} extended this binarization to activations as well. 
    TWN ~\cite{twn} proposed ternary quantization instead, increasing model expressiveness. Similarly, TTQ ~\cite{ttq} used ternary weights with a positive and negative scaling learned during training. 
    LR-Net ~\cite{lrnet} made use of both binary and ternary weights by using stochastic parameterization while INQ ~\cite{zhou2017incremental} constrained weights to powers of two and zero. FGQ ~\cite{ternaryfinegrained} categorized weights in different groups and used different scaling factors to minimize the element-wise distance between full and low-precision weights. \cite{wang2019haq} used the hardware accelerator's feedback to perform hardware-aware quantization using reinforcement learning. \cite{lqnets} jointly trained quantized networks and respective quantizers. \cite{reagen2017weightless} used Bloomier filters to compactly encode network weights.
    
    Similarly, quantization techniques can also be applied in the backward pass. Therefore, some previous work quantized not only weights and activations but also the gradients to augment training performance ~\cite{dorefa,pmlr-v37-gupta15,courbariaux2014training}. In particular, RQ ~\cite{rq}  propose a differentiable quantization procedure to allow for gradient-based optimization using discrete values and ~\cite{integerDNNs} recently proposed to discretize weights, activations, gradients, and errors both at training and inference time.
    
    These quantization techniques have great benefits and have shown to successfully reduce the computation requirements compared to full-precision models. However, all the aforementioned methods require re-training of the quantized network to achieve close to full-precision accuracy, which can introduce significant financial and environmental cost \cite{strubell2019energycost}.
    On the other hand, our method instantly quantizes pre-trained neural networks with minimal accuracy loss as compared to their full-precision counterparts \textit{without any kind of additional training}.

\section{Neural Networks and Monte Carlo Methods}\label{sec:NNMC}
    
    Neural networks make extensive use of randomization and random sampling techniques. Examples are random initialization of network weights, stochastic gradient descent~\cite{sgd}, regularization techniques such as Dropout~\cite{dropout} and DropConnect~\cite{dropconnect}, data augmentation and data shuffling, recurrent neural networks' regularization~\cite{merityRegularizing}, or the generator's noise input on generative adversarial networks~\cite{gans}.
    
    Many state-of-the-art networks use ReLU~\cite{nair2010rectified}, which has interesting properties such as scale-invariance. This enables a scaling factor to be propagated through all network layers without affecting the network's original output. This principle can be used to normalize network values, such as weights and activations, as further described in \Cref{sec:network_normalization}. After normalization, these values can be treated as probabilities, which enables the simulation of discrete probability densities to approximate the corresponding full-precision, continuous distributions (\Cref{sec:network_quantization}).
    
    \subsection{Network Normalization}\label{sec:network_normalization}
        
        Assuming the exclusive use of the ReLU activation function in the hidden layers, the scale-invariance property of the ReLU activation function allows for arbitrary scaling of the weights or activations without affecting the network's output. Given weights $w_{l-1,i,j}$ connecting the $i$-th neuron in layer $l-1$ to the $j$-th neuron in layer $l$, where $i \in \left[0, N_{l-1}-1\right]$ and $j \in \left[0, N_{l}-1\right]$, with $N_{l-1}$ and $N_l$ the number of neurons of layer $l-1$ and $l$, respectively. Let $a_{l,j}$ be the $j$-th activation in the $l$-th layer and $f \in \mathbb{R}^+$:
        $            a_{l,j} = max\Bigg\{0, \sum_{i=0}^{N_{l-1}-1}       w_{l-1,i,j} a_{l-1, i} + b_{l,j} \Bigg\} %\\
            = f \cdot max\Bigg\{0, \frac{\sum_{i=0}^{N_{l-1}-1}  w_{l-1,i,j} a_{l-1, i} + b_{l,j}}{f}\Bigg\}.$

        Biases and incoming weights for neuron $j$ in layer $l$ may then be normalized by $f = \|\mathbf{w}_{l-1,j}\|_1 = \sum_{i=0}^{N_{l-1}-1} |w_{l-1,i,j}|$,

        enabling weights to be seen as a probability distribution over all connections to a neuron. A similar procedure could be used to normalize all activations $a_{l,j}$ of layer $l$.
        
        Propagating these scaling factors forward layer by layer results in a single scalar (per output), which converts the outputs of the normalized network to the same range as the original network. This technique allows for the usage of integer weights and activations throughout the entire network without requiring rescaling or conversion to floating point at every layer.

    \subsection{Network Quantization}\label{sec:network_quantization}

        Taking advantage of the normalized network, we can simulate discrete probability densities by constructing a probability density function (PDF) and then sampling from the corresponding cumulative density function (CDF). The number of references of a weight is then the quantized integer approximation of the continuous value. For simplicity, the following discussion shows the quantization procedure for weights; activations can be quantized in the same way at inference time. 
        
        Without loss of generality, given $n$ weights, assuming $\sum_{k=0}^{n-1}|w_k| = \|w\|_1 = 1$ and defining a partition of the unit interval by $P_m := \sum_{k=1}^{m}|w_k|$ we have the following partitions:
        
        \begin{equation}
	        \begin{tikzpicture}
	        \draw [densely dashed,thick] (0,0) -- (5,0);
	        \draw [thick] (0,0) -- (2.7,0);
	        \draw [thick] (3.7,0) -- (5,0);
	        \draw [thick] (0,-0.05) -- (0,0.05);
	        \node [below] at (-0.2,0) {$0 = P_0$};
	        \draw [thick] (5,-0.05) -- (5,0.05);
	        \node [below] at (5.7,0) {$P_{n-1} = 1$};
	        \draw [thick] (1,-0.05) -- (1,0.05);
	        \node [below] at (1,0) {$P_1$};
	        \draw [thick] (2.5,-0.05) -- (2.5,0.05);
	        \node [below] at (2.5,0) {$P_2$};
	        \draw [thick] (4.2,-0.05) -- (4.2,0.05);
	        \node [below] at (4.2,0) {$P_{n-2}$};
	        \node [above] at (0.5,0) {$|w_1|$};
	        \node [above] at (1.75,0) {$|w_2|$};
	        \node [above] at (4.6,0) {$|w_{n-1}|$};
	        \end{tikzpicture}
        \label{eq:partitions_fig}
        \end{equation}
        
        Then, given $N$ uniformly distributed samples $x_i \in [0,1)$, we can approximate the weight distribution as follows:
        
        \begin{equation}
            \sum_{j=0}^{n-1}w_j a_j \approx \frac{1}{N} \sum_{i=0}^{N-1} \underbrace{\text{sign}(w_{j_i})}_{\in \{-1, 0, 1\}}\times a_{j_i},
            \label{eq:approx}
        \end{equation}
        where $j_i \in \{0, \ldots, n-1\}$ is uniquely determined by $P_{j_i-1} \leq x_i < P_{j_i}$.
        
        One can further improve this sampling process by using \textit{jittered equidistant sampling}. Thus, given a random variable $\xi \in [0,1)$, we generate N uniformly distributed samples $x_i \in [0, 1)$ such that $x_i = \dfrac{i+\xi}{N}$,
        where $i \in \{0, \ldots, N-1\}$. The combination of equidistant samples and a random offset improves the weight approximation, as the samples are more uniformly distributed. The variance of different sampling seeds is discussed in the Appendix.

    \section{Monte Carlo Quantization (MCQ)}\label{sec:MCQ}
    
        Our approach builds on the aforementioned ideas of network normalization and quantization using random sampling to quantize an entire pre-trained full-precision neural network. As before, we focus on weight quantization; online activation quantization is discussed in \Cref{sec:activation_quantization}. Our method, called Monte Carlo Quantization (MCQ), consists of the following steps, which are executed layer by layer:
        
        \begin{enumerate}[label=(\arabic*)]
          \item Create a probability density function (PDF) for all $N_{l,w}$ weights 
          of layer $l$ such that $\sum_{i=0}^{N_{l,w}-1}|w_{l,i}| 
          = 1$  (\Cref{sec:method_normalization}).
          
          \item Perform importance sampling on the weights based on their magnitude by sampling from the corresponding cumulative density function (CDF) and counting the number of hits per weight (\Cref{sec:method_sampling}).
          
          \item Replace each weight with its quantized integer value, \textit{i.e.} its hit count, to obtain a low bit-width, integer weight representation (\Cref{sec:method_quantization}).
        \end{enumerate}

        The pseudo-code for our method is shown in \Cref{alg:method} of the Appendix.
        \Cref{fig:method} illustrates both the normalization and importance sampling processes for a layer with 10 weights and 1 sample per weight, \textit{i.e.} $K = 1.0$.
        
        \begin{figure*}[ht] 
            \centering
            \begin{subfigure}{.24\linewidth}
               \centering \includegraphics[width=\linewidth]{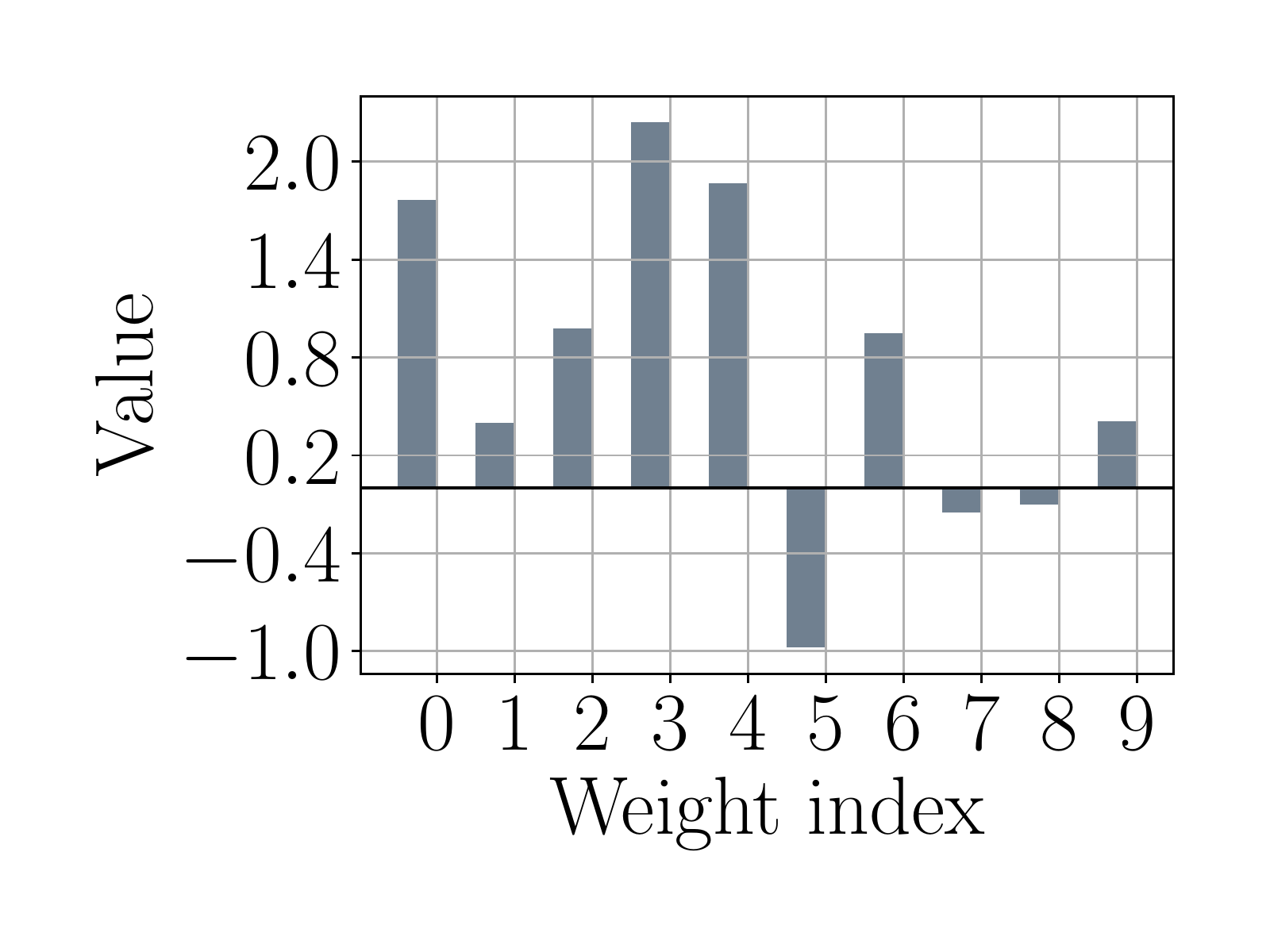} 
               \vspace{-2\baselineskip}
               \caption{Full-precision weights}
            \end{subfigure}
            \begin{subfigure}{.24\linewidth}
                \centering
                \includegraphics[width=\linewidth]{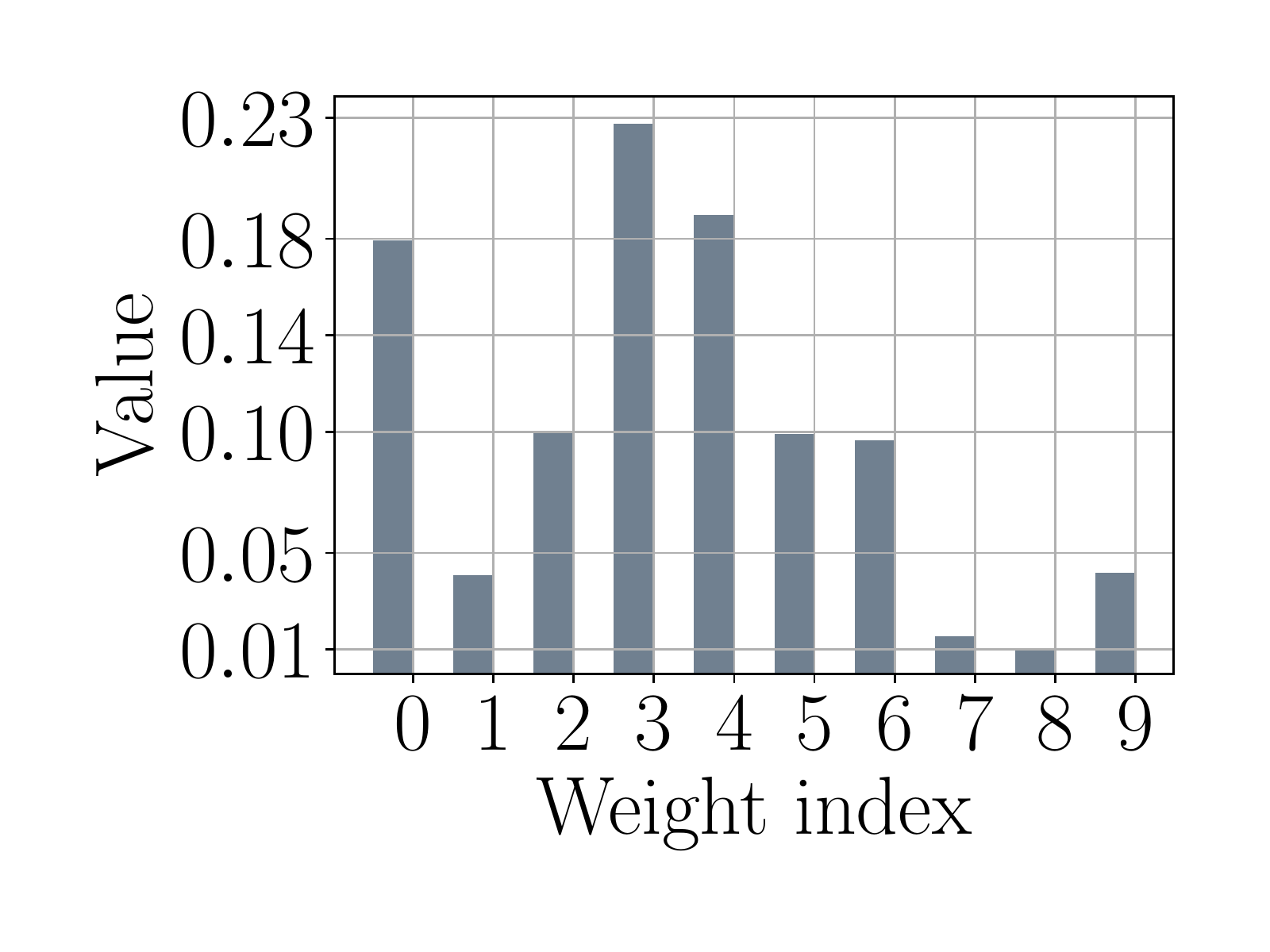}
                \vspace{-2\baselineskip}
                \caption{PDF}
            \end{subfigure}
            \begin{subfigure}{.24\linewidth}
                \centering
                \includegraphics[width=\linewidth]{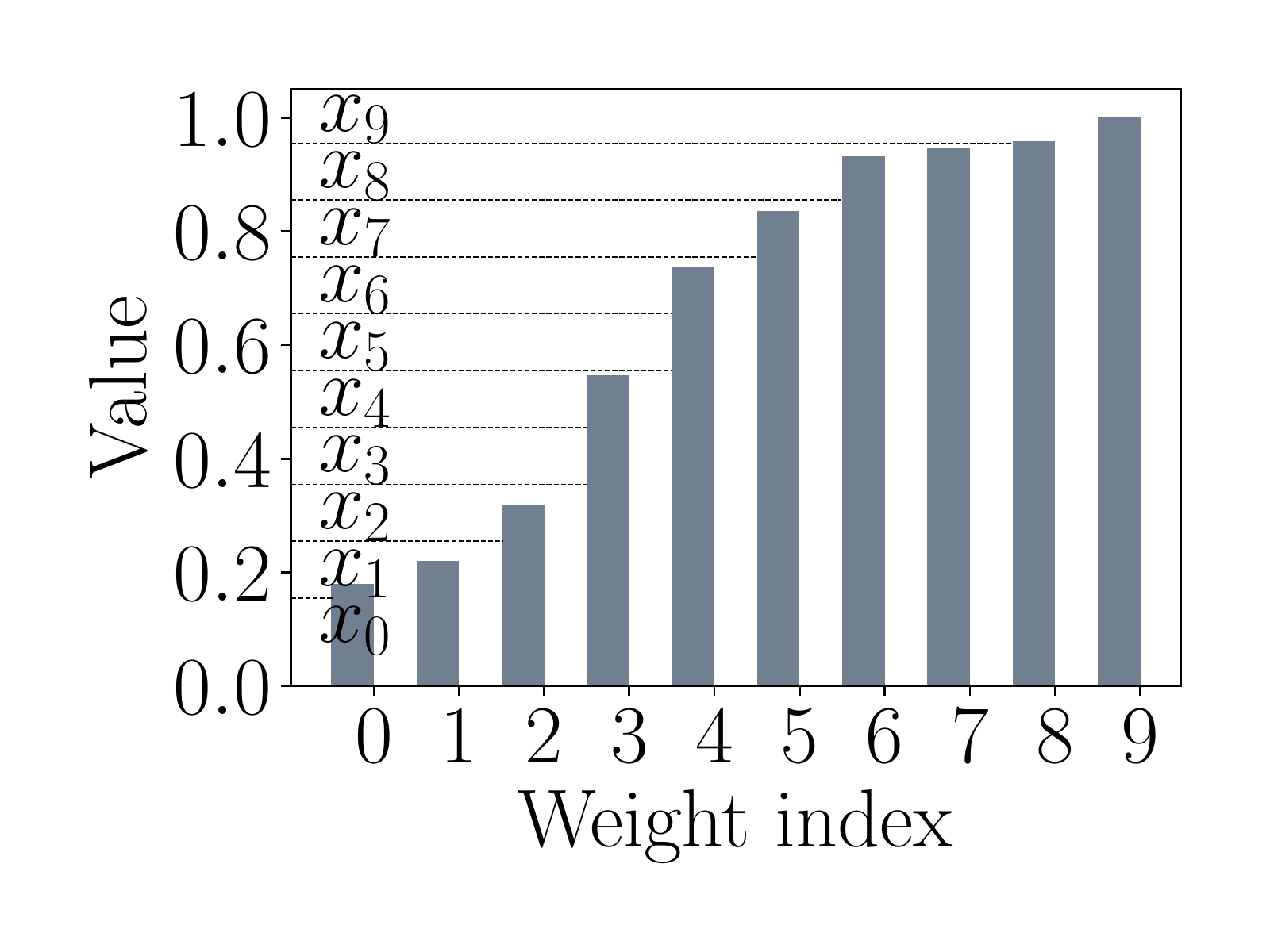} 
                \vspace{-2\baselineskip}
                \caption{Sampling on CDF}
            \end{subfigure}
            \begin{subfigure}{.24\linewidth}
                \centering
                \includegraphics[width=\linewidth]{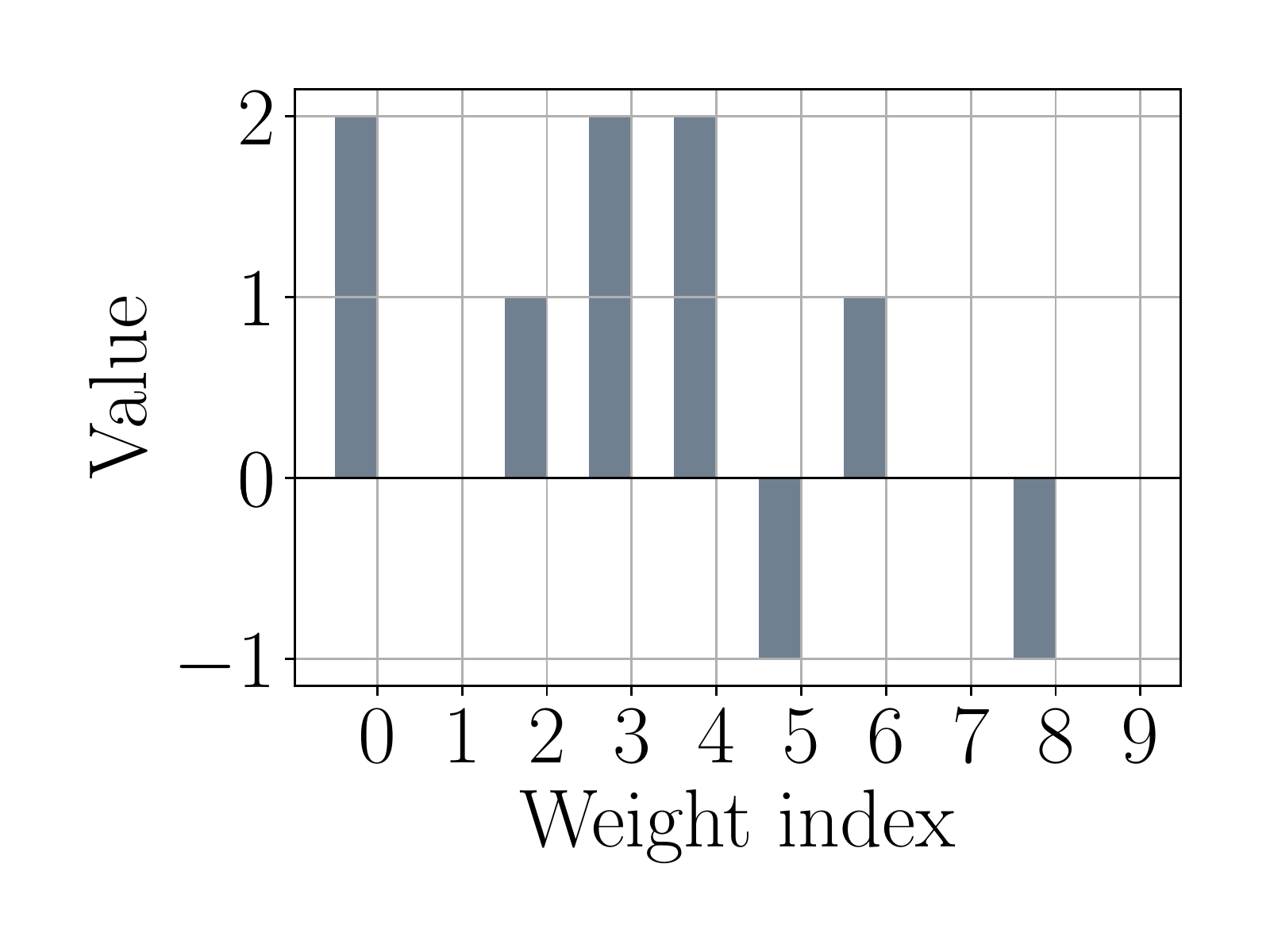} 
                \vspace{-2\baselineskip}
                \caption{Integer weights}
            \end{subfigure}
            \caption{Starting from full-precision weights (a), we create a PDF of the sorted absolute values (b) and uniformly sample from the corresponding CDF (c). The sampling process produces quantized integer network weights based on the number of hits per weight (d). Note that since weights 7, 8, and 9 were not hit, sparsity is introduced which can be exploited by hardware accelerators. }
            \label{fig:method}
        \end{figure*}
    
    \subsection{Layer Normalization}\label{sec:method_normalization}
        
        %layer-wise
        Performing normalization neuron-wise, as introduced in \Cref{sec:network_normalization} may result in an inferior approximation, especially when the number of weights to sample from is small, as for example in convolutional layers with a small number of filters or input channels. 
        To mitigate this, we propose to normalize all neurons simultaneously in a layer-wise manner. This has the additional advantage that samples can be redistributed from low-importance neurons to high-importance neurons (according to some metric), resulting in an increased level of sparsity. Additionally, there is more opportunity for global optimization, so the overall weight distribution approximation improves as well.
        
        %updated scaling factor
        We use the 1-norm of all weights of a given layer $l$ as the scaling factor $f$ used to perform weight normalization. Thus, each normalized weight can be seen as a probability with respect to all connections between layer $l-1$ and layer $l$, instead of a single neuron. This layer-wise normalization technique is similar to Weight Normalization~\cite{weightnormalization}, which decouples the neuron weight vector magnitude from its direction.

    \subsection{Importance Sampling}\label{sec:method_sampling}
    
        As introduced in \Cref{sec:network_quantization}, we generate ternary samples (hit positive weight, hit negative weight, or no hit), and count such hits during the sampling process. Note that even though the individual samples are ternary, the final quantized values may not be, because a single weight can be sampled multiple times.
        We use jittered equidistant (stratified) sampling, to ensure uniform distribution. Given a random variable $\xi \in [0,1)$, we generate N samples $x_i \in [0, 1)$ such that $x_i = \dfrac{i+\xi}{N}$, where $i \in \{0, \ldots, N-1\}$.  This stratified sampling strategy also reduces the cost of the sampling process from $\mathcal{O}(N log N)$ to $\mathcal{O}(N)$, as searching for the value corresponding to a sample does requires a binary search as it would for fully random sampling.
        The number of samples $N=K \cdot N_{values}$, where $K \in \mathbb{R}^+$ is a user-specified parameter to control the number of samples and $N_{values}$ represents the number of weights of a given layer.
        By varying K, the computational cost of sampling can be traded off better approximation (more bits per weight) of the original weight distribution, leading to higher accuracy. In our experiments, $K$ is set the same for all network layers.
    
        One simple modification to enhance the quality of the discrete approximation is to sort the continuous values prior to creating the PDF. Applying sorting mechanisms to Monte Carlo schemes has been shown to be beneficial in the past~\cite{l2008randomized, l2018sorting}. Sorting groups smaller values together in the overall distribution. Since we are using a uniform sampling strategy, smaller weights are then sampled less often, which results in both higher sparsity and a better quantized approximation of the larger weights in practice. This effect is particularly significant on smaller layers with fewer weights.

        Since the quantized integer weights span a different range of values than the original weights, and biases remain unchanged, care must be taken to ensure the activations of each neuron are calculated correctly. After the integer multiply-accumulate (MAC) operation, the result must then be scaled by $\frac{f}{N}$ before adding the bias. This requires the storage of one floating point scaling value per layer. However, weights are stored as low bit-width integers and the computational cost is greatly reduced since the MAC operations use low-precision integers only instead of floating point numbers.
  
    \subsection{Layer Quantization}\label{sec:method_quantization}
    
        The number of bits required for the weights $B_{W_l} \in \mathbb{N}$, for layer $l$ and its quantized weights $Q(w_{l,i})$, corresponds to the bit amount needed to represent the highest hit count during sampling, including its sign: $B_{W_l} = 1 + \left\lfloor \log_2\left(
            \max_{0 \leq i \leq N_w-1}|Q(w_{l,i})|\right)\right\rfloor + 1 $.
        Alternatively, positive and negative weights could be separated into two sets.
      
    \subsection{Online Quantization} \label{sec:activation_quantization}
        
        While weights are quantized offline, \textit{i.e.} after training and before inference, activations are quantized online during inference time using the same procedure as weight quantization previously described. Thus, in the normalization step (Section \ref{sec:method_normalization}), all $N_{l,a}$ activations of a given layer $l$ are treated as a probability distribution over the output features, such that $\sum_{j=0}^{N_{l,a}-1}|a_{l,j}| = 1$.
        Then, in the importance sampling step (\Cref{sec:method_sampling}), activations are sub-sampled using possibly different relative sampling amounts, \textit{i.e.} $K$, than the ones used for the weights (we use the same $K$ for both weights and activations in all of our experiments).
        The required number of bits $B_{A_l}$ for the quantized activations $Q(a_{l,j})$ can also be calculated similarly as described in \Cref{sec:method_quantization}, although no additional bit sign is required when using ReLU since all activations are non-negative.

%%%%%%%%% BODY TEXT
\section{Experiments}
    \label{sec:results}
    
    The proposed method is extensively evaluated on a variety of tasks: for image classification we use CIFAR-10~\cite{cifar}, SVHN~\cite{svhn}, and ImageNet~\cite{imagenet}, on multiple models each. We further evaluate MCQ on language modeling, speech recognition, and machine translation, to assess the preformance of MCQ across different task domains.
    
    Due to the automatic quantization done by MCQ, some layers may be quantized to lower or higher levels than others. We indicate the quantization level for the whole network by the average number of bits, \textit{e.g.} '8w-32a' means that on average 8 bits were used for weights and 32 bits for activations on each layer.
    
    Many works note that quantizing the first or last network layer reduces accuracy significantly \cite{deepcompression, dorefa, twn}. We use footnotes \footnote{\label{not_first_layer}Not quantizing weights in the first layer.}, \footnote{\label{not_last_layer}Not quantizing weights in the last layer.}, and \footnote{\label{first_layer_special}Using higher precision (8w-8a) for the first layer.} to denote the special treatment of first or last layers respectively. For MCQ we report the results with both quantized and full-precision first layer. 
    We do not quantize Batch Normalization layers as the parameters are fixed after training and can be incorporated into the weights and biases~\cite{integerDNNs}. 
    
    \Cref{tab:results_cifar,tab:results_svhn,tab:results_imagenet,tab:results_nlp} show the accuracy difference $\Delta$ between the quantized and full-precision models. For other compared works this difference is calculated using the baseline models reported in each of the respective works. We didn't perform any search over random sampling seeds for MCQ's results.
\subsection{CIFAR-10}
    
    The best accuracies on VGG-7, VGG-14, and ResNet-20 produced by our method using $K= 1.0$ on CIFAR-10 are shown in \Cref{tab:results_cifar}. We refer to the Appendix for model and training details. MCQ outperforms or shows competitive results showing minimal accuracy loss on all tested models against the compared methods that require network re-training. The full-precision baselines for BNN~\cite{bnn} and XNOR-Net~\cite{xnornet} are from BC~\cite{courbariaux2015binaryconnect} as these works use the same model. Similarly, BWN~\cite{xnornet}'s results on VGG-7 are the ones reported in TWN~\cite{twn} since they did not report the baseline in the original paper.
    
\begin{figure*}[ht!] 
      \begin{subfigure}{.305\linewidth}
        \includegraphics[width=\linewidth]{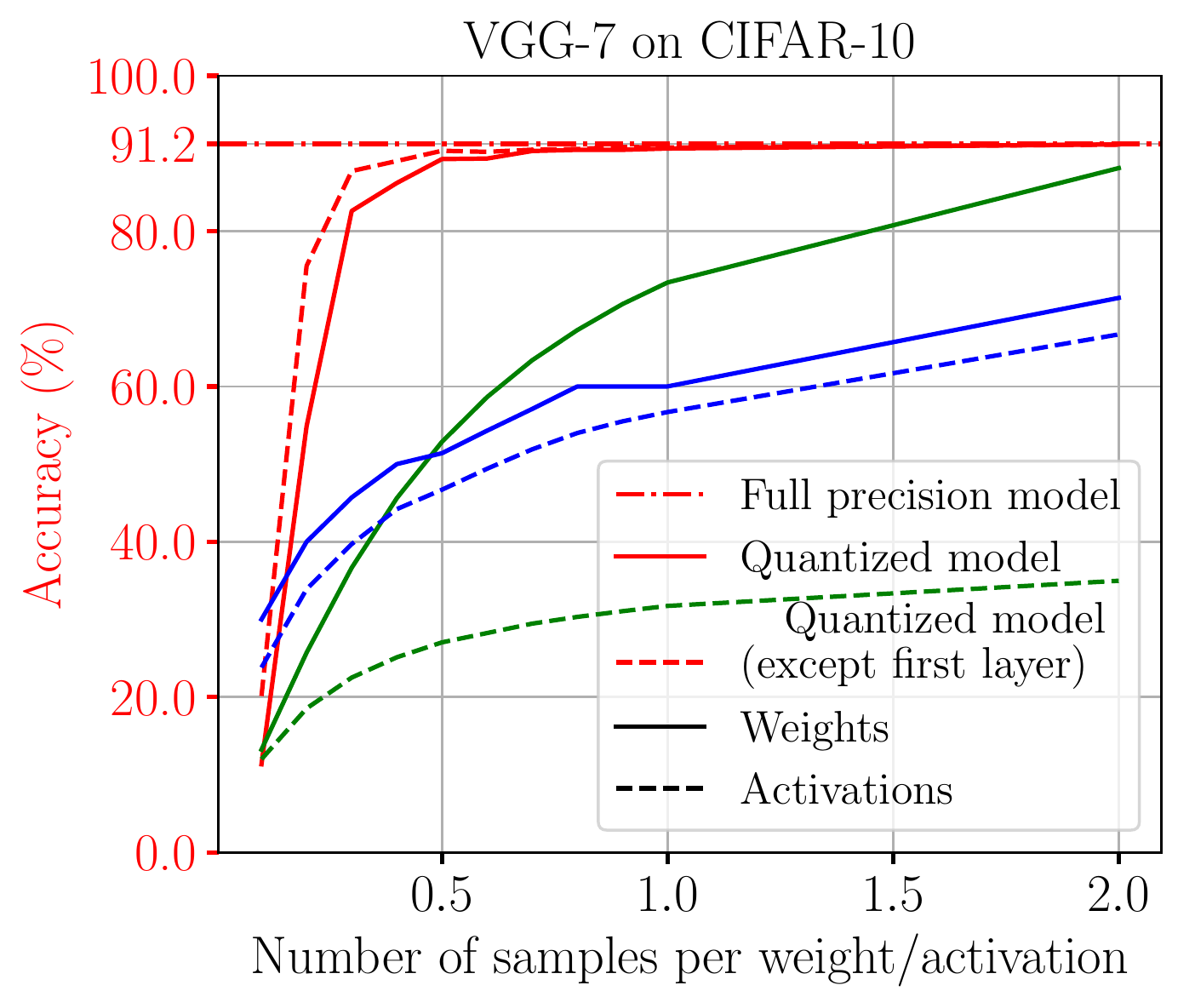} 
      \end{subfigure}
      \begin{subfigure}{.305\linewidth}
        \includegraphics[width=\linewidth]{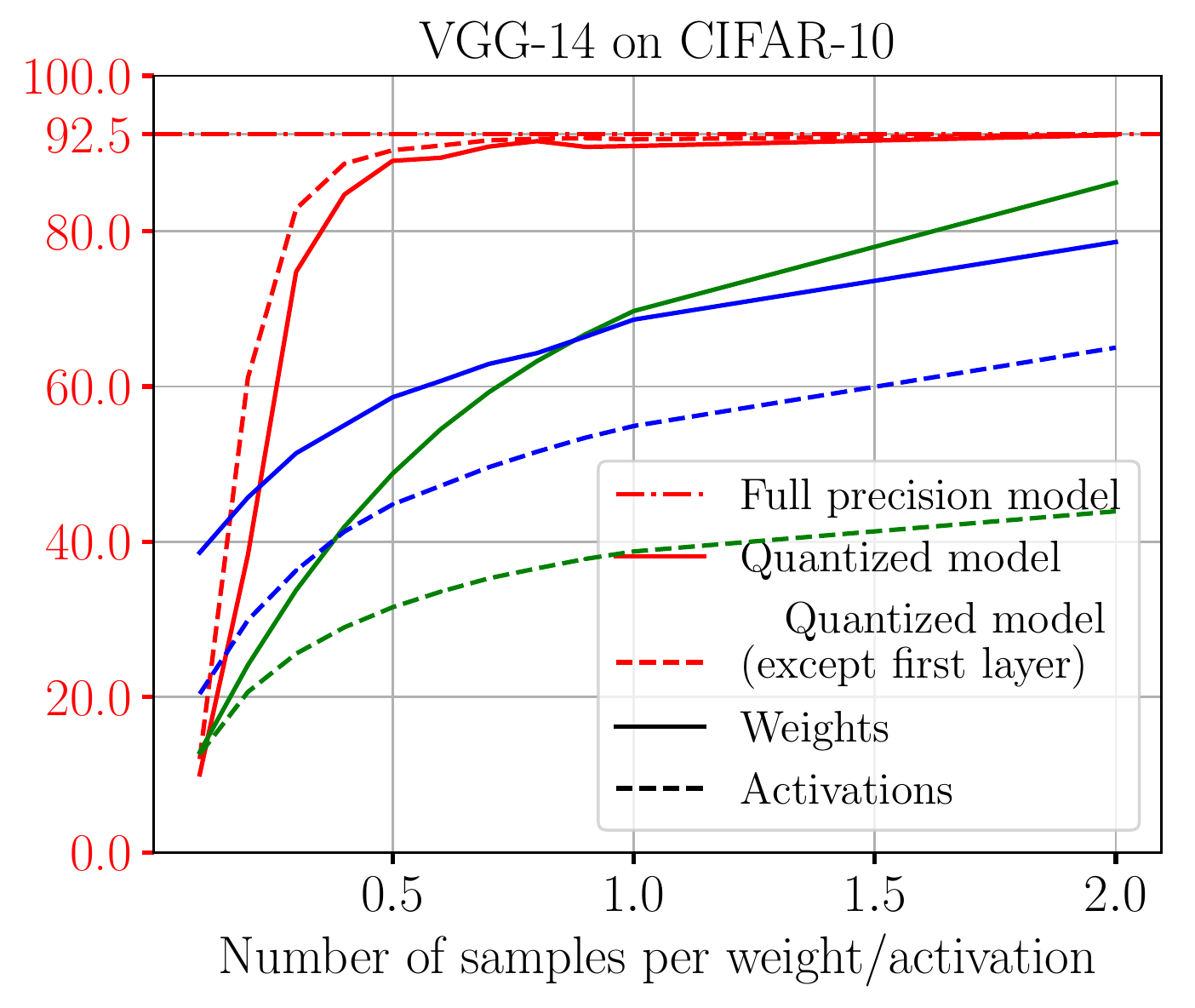} 
      \end{subfigure}
      \begin{subfigure}{.305\linewidth}
        \includegraphics[width=\linewidth]{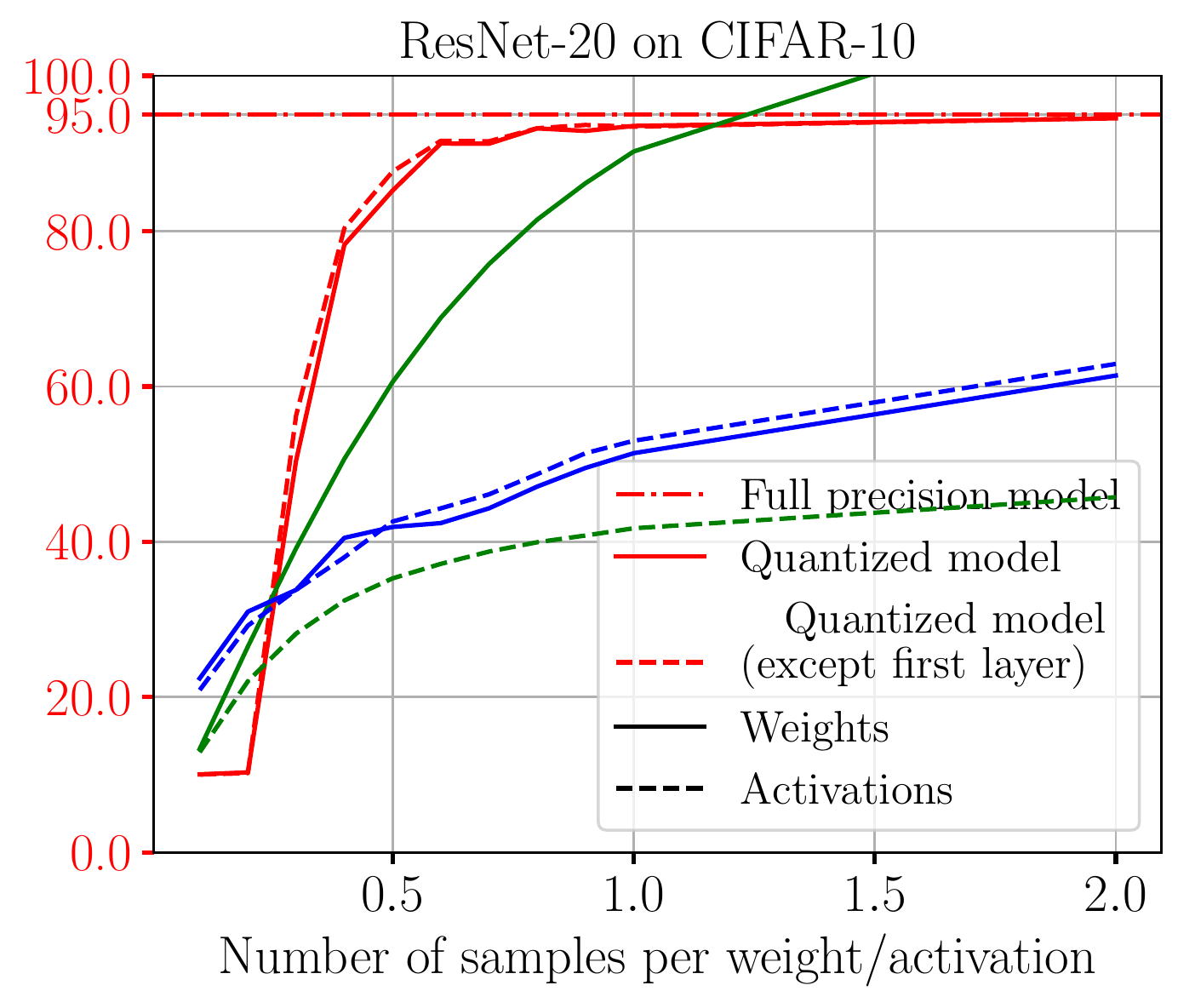}
      \end{subfigure}
       \begin{subfigure}{.062\linewidth}
        \includegraphics[width=\linewidth]{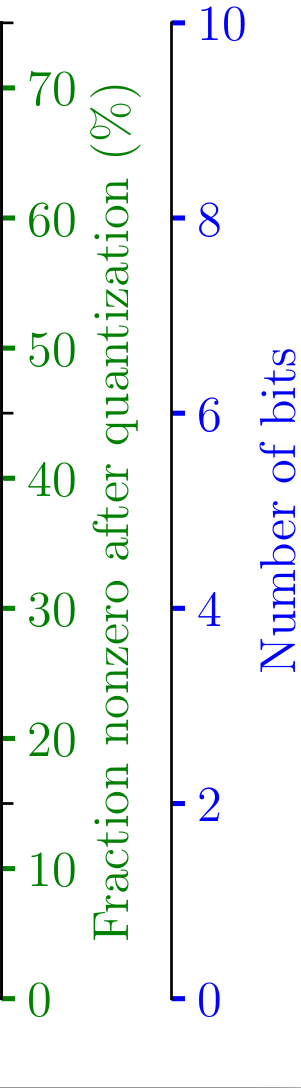}
      \end{subfigure}
      \caption{Results of quantizing both weights and activations on CIFAR-10 using different sampling amounts. The quantized models reach close to full-precision accuracy at around half the sample size while using only around half the weights and one-third of the activations of the full-precision models.} 
      \label{fig:cifar10}
    \end{figure*}

    \Cref{fig:cifar10} shows the effects of varying the amount of sampling, \textit{i.e.} using $K \in \left[0.1 ... 2.0\right]$.
    The average percentage of used weights/activations per layer and corresponding bit-widths of the final quantized model is also presented on each graph. We observe a rapid increase of the accuracy even when sparsity levels are high on all tested models.
    
    \begin{table*}[ht!]
        \caption{Accuracy results on CIFAR-10 when quantizing either weights or activations or both. Quantizing only the weights leads to an accuracy loss of $\approx1.0\%$ in the worst case. Quantizing both weights and activations does not reduce accuracy on VGG-7 while ResNet-20's accuracy decreases by $\approx1.0\%$. Quantizing the first layer results in an additional $\approx0.5\%$ accuracy loss on all models.}
        \label{tab:cifar10}
        \begin{adjustbox}{width=1.0\textwidth}
        \centering
        \begin{small}
        \begin{sc}
        \begin{tabular}{lcccr}
            \toprule
            Method & VGG-7 & VGG-14 & ResNet-20\\
            \midrule
            Full Precision (32w-32a) & 91.23 & 92.49 & 95.02\\
            $\Delta$ MCQ (quantized w) & -0.48 (6.1w-32a) / +0.04\footref{not_first_layer} (6.1w-32a) & -1.04 (6.7w-32a) / -0.50\footref{not_first_layer} (6.8w-32a) & -0.84 (5.1w-32a) / -0.54\footref{not_first_layer} (5.1w-32a) \\
            
            $\Delta$ MCQ (quantized a) & -0.12\footref{not_first_layer} (32w-5.68a)
            & -0.06\footref{not_first_layer}(32w-5.51a) 
            & -0.28\footref{not_first_layer}(32w-6.3a) \\
            
            $\Delta$ MCQ (quantized w + a) & -0.58 (6.1w-5.6a) / -0.13\footref{not_first_layer} (6.1w-5.6a)  & -1.08 (6.6w-5.3a) / -0.54\footref{not_first_layer} (6.8w-5.5a) & -1.77 (5.1w-5.3a) / -1.21\footref{not_first_layer} (5.1w-5.3a)\\
            \midrule
            $\Delta$ TTQ (2w-32a) & - & - & -0.64\footref{not_first_layer} \\
            $\Delta$ dLAC (2w-32a) & - & -3.0 / -1.4\footref{not_first_layer} & - \\
            $\Delta$ TWNs (2w-32a) & -0.06 & - & - \\
            $\Delta$ BC (1w-32a) & +0.74 & - & - \\
            $\Delta$ BNN (1w-1a) & +0.49\footref{not_first_layer} & - & - \\ 
            $\Delta$ BWN (1w-32a) & -0.36 / +0.76\footref{not_first_layer}  & - & - \\ 
            $\Delta$ XNOR-Net (1w-1a) & +0.47\footref{not_first_layer} & - & - \\ 
            $\Delta$ RQ (8w-8a)) & +0.25 & - & - \\ 
            $\Delta$ LR-net (2w-32a) & -0.11\footref{not_last_layer} & - & - \\ 
            \bottomrule
        \end{tabular}
        \end{sc}
        \end{small}
        \end{adjustbox}
        \label{tab:results_cifar}
        \end{table*}
    
\subsection{SVHN}
    
    For SVHN, the tested models are identical to the compared methods. Models B, C, and D have the same architecture as Model A but with a 50\%, 75\%, and 87.5\% reduction in the number of filters in each convolutional layer, respectively \cite{dorefa}. We refer to the Appendix for further details.
    
\begin{figure*}[t] 
        \begin{subfigure}{.31\linewidth}
            \includegraphics[width=\linewidth]{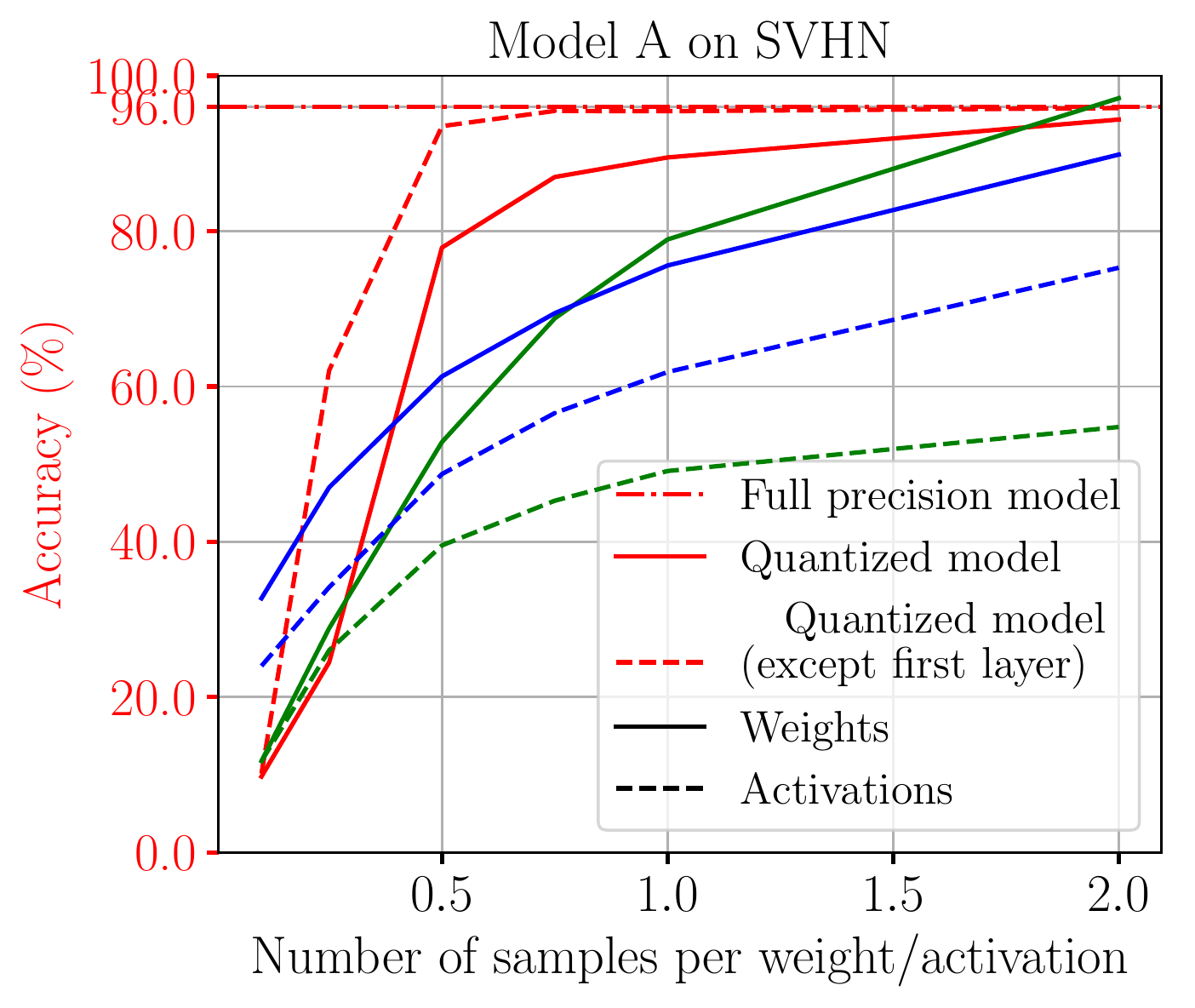}
        \end{subfigure}
        \begin{subfigure}{.31\linewidth}
            \includegraphics[width=\linewidth]{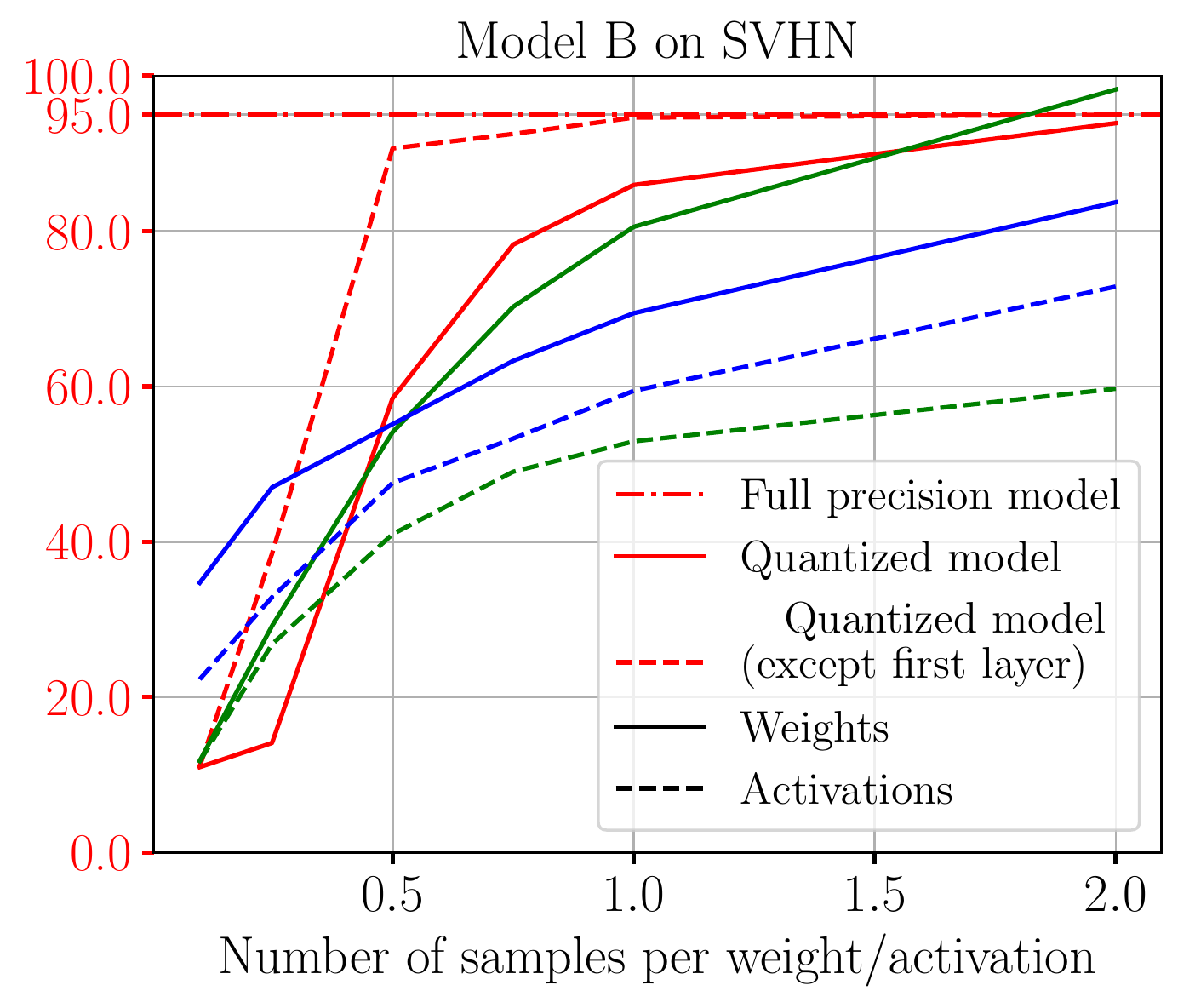}
        \end{subfigure}
        \begin{subfigure}{.31\linewidth}
            \includegraphics[width=\linewidth]{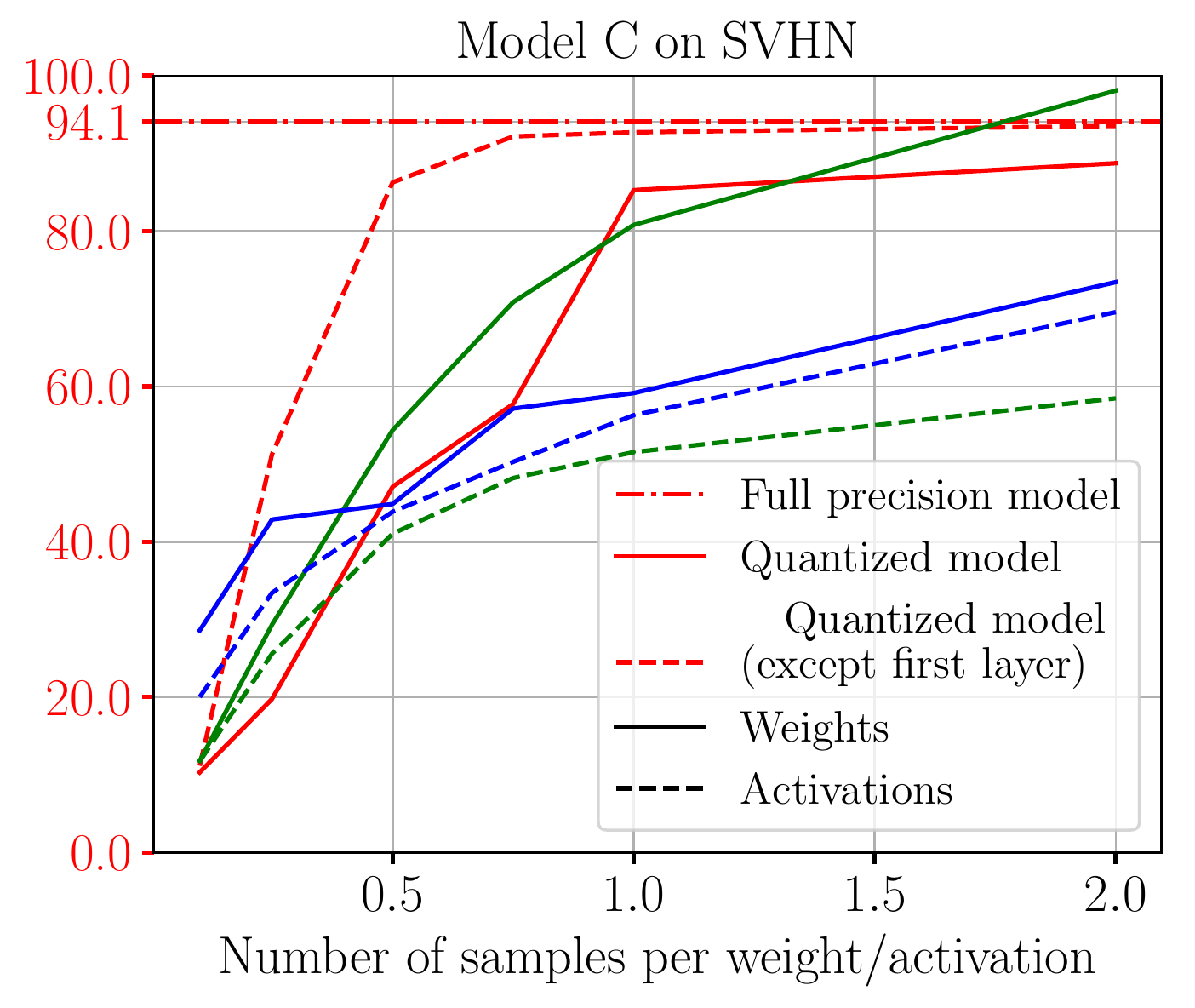}
        \end{subfigure}
        \begin{subfigure}{.055\linewidth}
            \includegraphics[width=\linewidth]{Images/svhn/weights+activations/rightax}
        \end{subfigure}
        
        \begin{center}
            \begin{subfigure}{.40\linewidth}
                \includegraphics[width=\linewidth]{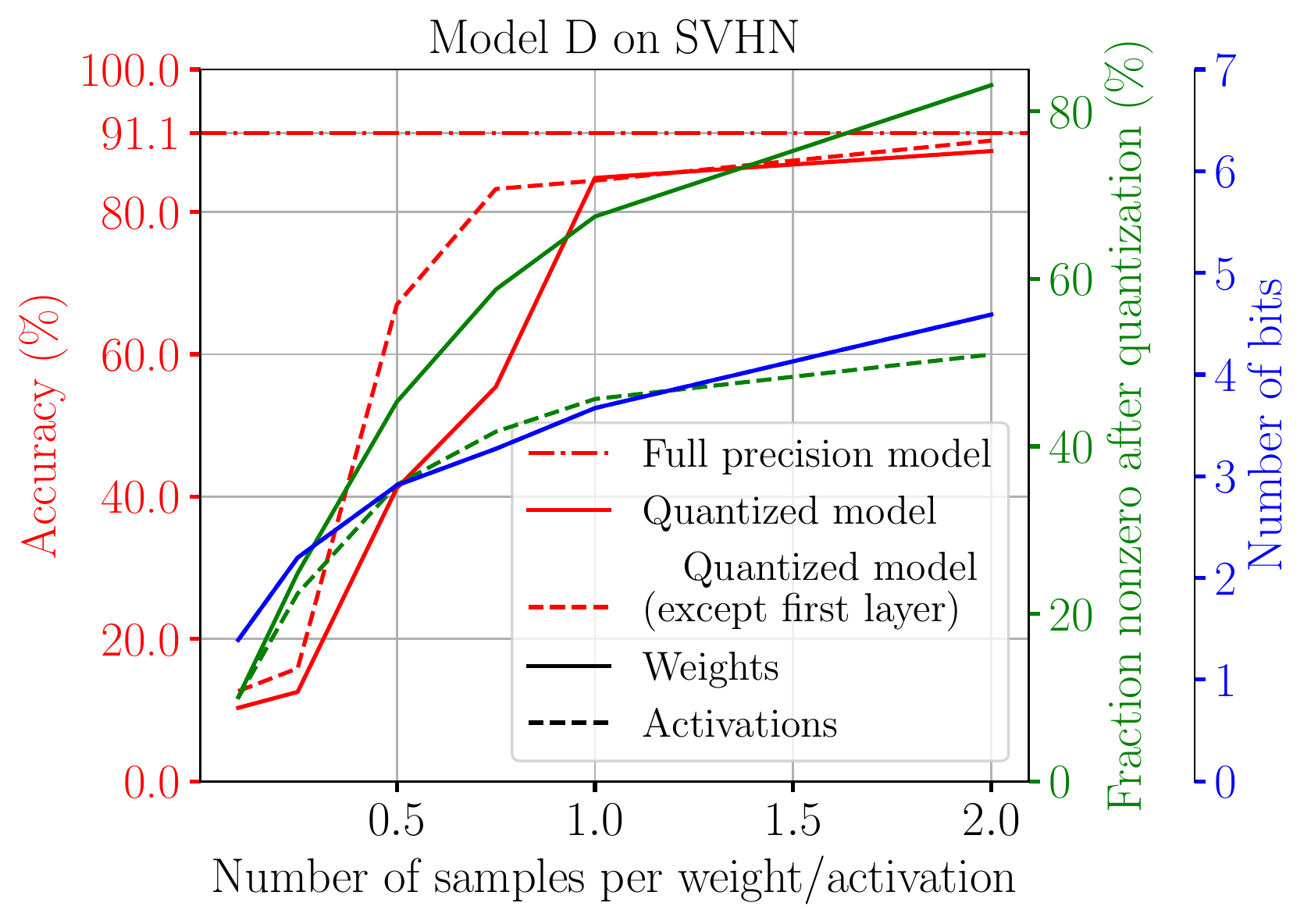}
            \end{subfigure}
            \begin{subfigure}{.40\linewidth}
                \includegraphics[width=\linewidth]{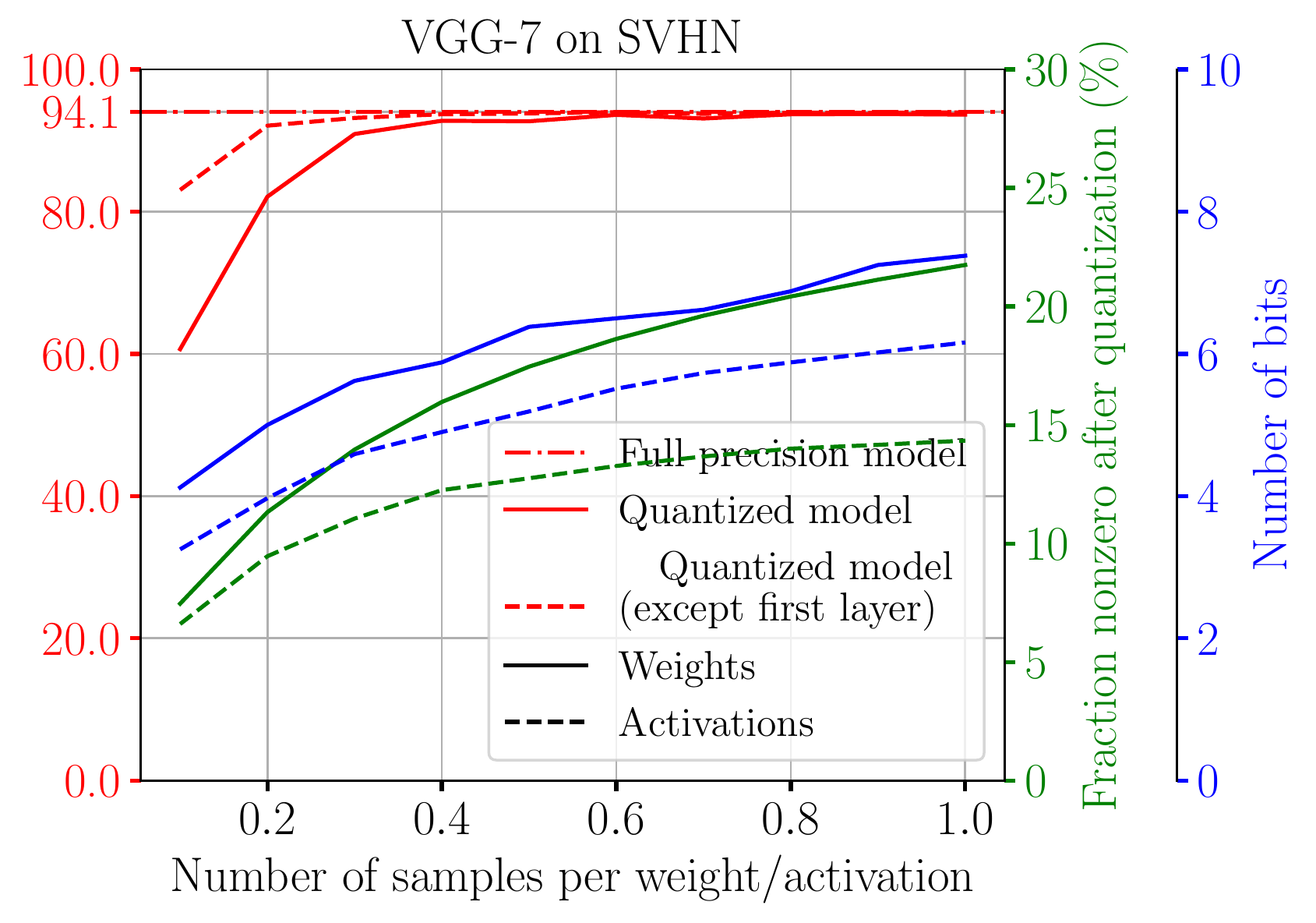}
            \end{subfigure}
        \end{center}
        \caption{Results of quantizing both weights and activations on SVHN using different sampling amounts. The quantized VGG-7* model reaches close to full-precision accuracy using around 0.5 samples per weight/activation, requiring around 8 bits and using 22\% of the weights of the original model, with 22\% nonzero activations. Model A, B, C, and D are less redundant models that require more sampling to achieve close to full-precision accuracy.
        } 
        \label{fig:svhn}
    \end{figure*}

    \Cref{tab:results_svhn} shows MCQ's results for several models on SVHN using $K = 1.0$. On bigger models, \textit{i.e.} VGG-7* and Model A, we see minimal accuracy loss when compared to the full-precision baselines. For the smaller models, we observe a slight accuracy degradation as model size decreases due to the reduction in the sample size, resulting in a poorer approximation. However, we used only about 4 bits per weight/activation for such models. Thus, increasing the number of samples would improve accuracy while still maintaining a low bit-width.
    \Cref{fig:svhn} illustrates the consequences of varying the number of samples. Less samples are required than on CIFAR-10 for bigger models to achieve close to full-precision accuracy. Potentially this is because layers have a larger number of weights and activations, so a larger sample size reduces quantization noise since the important values being more likely to be better approximated. 
    
    \begin{table*}[t]
        \caption{Accuracy results on SVHN when quantizing weights, activations, or both. On VGG-7*, MCQ shows minimal accuracy loss when quantizing both weights and activations and close to no accuracy loss when not quantizing the first layer. For models A, B, C, and D the accuracy lowers as the model size decreases. Quantizing only the activations barely lowers the baseline accuracy.}
        \label{tab:svhn}
        \begin{adjustbox}{width=1.0\textwidth}
        \centering
        \begin{small}
        \begin{sc}
        \begin{tabular}{lcccccr}
            \toprule
            Method & VGG-7* & Model A & Model B & Model C & Model D\\
            \midrule
            Full Precision (32w-32a) & 94.06 & 96.01 & 95.03 & 94.08 & 91.08 \\
            $\Delta$ MCQ (quantized w) & -0.30 (7.3w-32a) / -0.02\footref{not_first_layer} (7.0w-32a) 
            & -0.20\footref{not_first_layer} (5.1w-32a) 
            & -0.30\footref{not_first_layer} (4.8w-32a) 
            & -1.48\footref{not_first_layer} (4.1w-32a) 
            & -2.17\footref{not_first_layer} (4.1w-32a)\\
            
            $\Delta$ MCQ (quantized a) & -0.04 (32w-7.15a)  
            & +0.01\footref{not_first_layer} (32w-5.28a)
            & -0.03\footref{not_first_layer} (32w-5.11a) 
            & -0.12\footref{not_first_layer} (32w-4.88a) 
            & -0.11\footref{not_first_layer} (32w-4.58a)\\
            
            $\Delta$ MCQ (quantized w + a) & -0.32 (7.2w-6.0a) / -0.06\footref{not_first_layer} (7.0w-5.5a)  
            & -0.40\footref{not_first_layer} (5.1w-4.2a) 
            & -0.56\footref{not_first_layer} (4.8w-4.1a) 
            & -2.13\footref{not_first_layer} (4.1w-3.9a) 
            & -3.72\footref{not_first_layer} (4.1w-3.7a)\\
            
            \midrule
            $\Delta$ DoReFa (1w-1a)  & - & -0.4\footref{not_first_layer}$^,$\footref{not_last_layer} & -1.2\footref{not_first_layer}$^,$\footref{not_last_layer} & -5.1\footref{not_first_layer}$^,$\footref{not_last_layer} & -10.9\footref{not_first_layer}$^,$\footref{not_last_layer} \\
            $\Delta$ BC (1w-32a)  & +0.14 & - & - & - & - \\
            $\Delta$ BNN (1w-1a)  & -0.09\footref{not_first_layer} & - & - & - & - \\
            \bottomrule
        \end{tabular}
        \end{sc}
        \end{small}
        \end{adjustbox}
        \label{tab:results_svhn}
    \end{table*}
    
    \subsection{ImageNet}
    
    For ImageNet, we evaluate MCQ on AlexNet, ResNet-18, and ResNet-50 using the pre-trained models provided by Pytorch's model zoo~\cite{pytorch}). \Cref{tab:imagenet} shows the results on ImageNet with $K = 5.0$ for the different models. The results shown for DoReFa, BWN, TWN~\cite{dorefa,xnornet,twn} are the ones reported in TTQ~\cite{ttq}.
    
\begin{figure*}[ht] 
        \begin{subfigure}{.305\linewidth}
            \includegraphics[width=\linewidth]{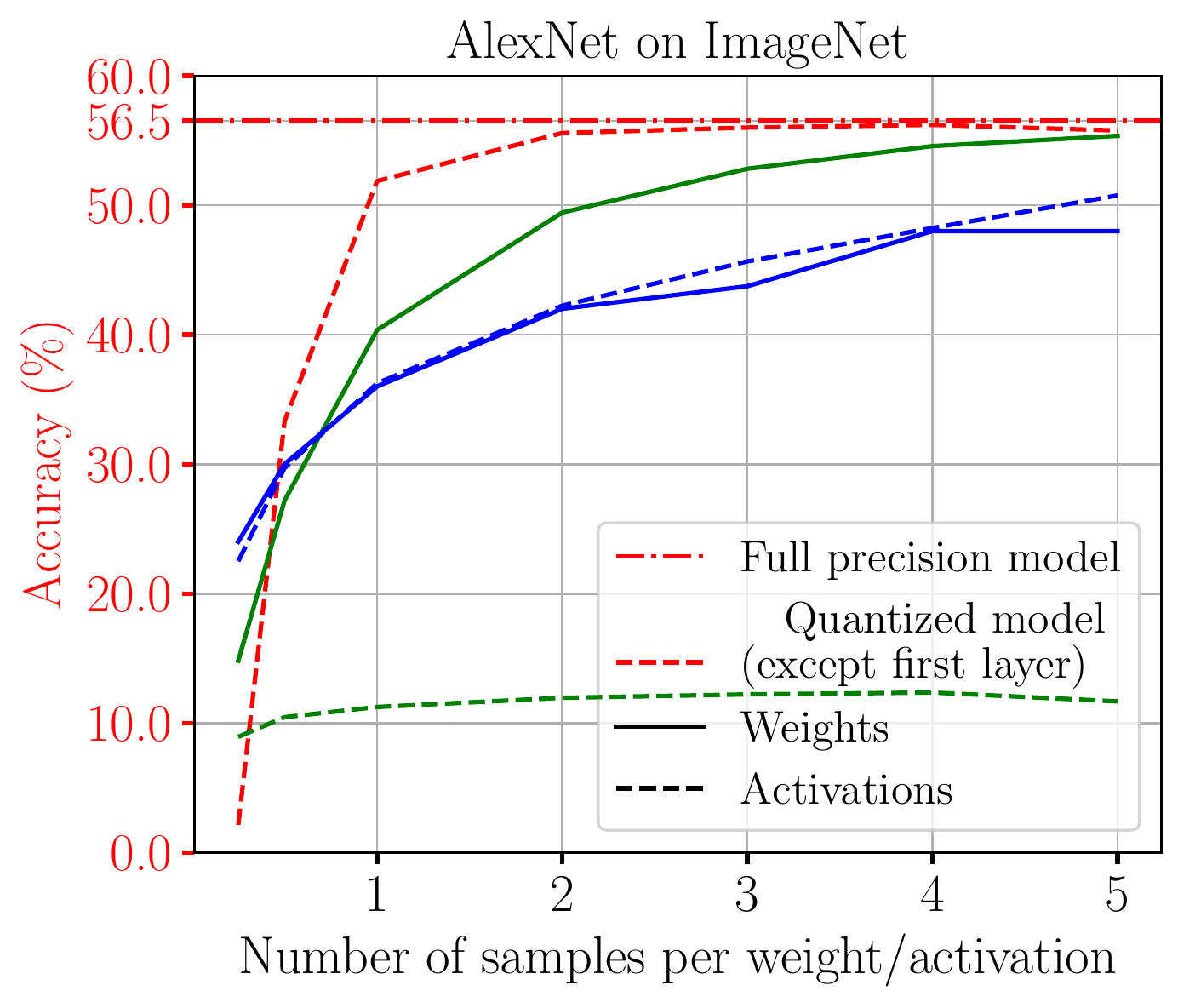}
        \end{subfigure}
        \begin{subfigure}{.305\linewidth}
            \includegraphics[width=\linewidth]{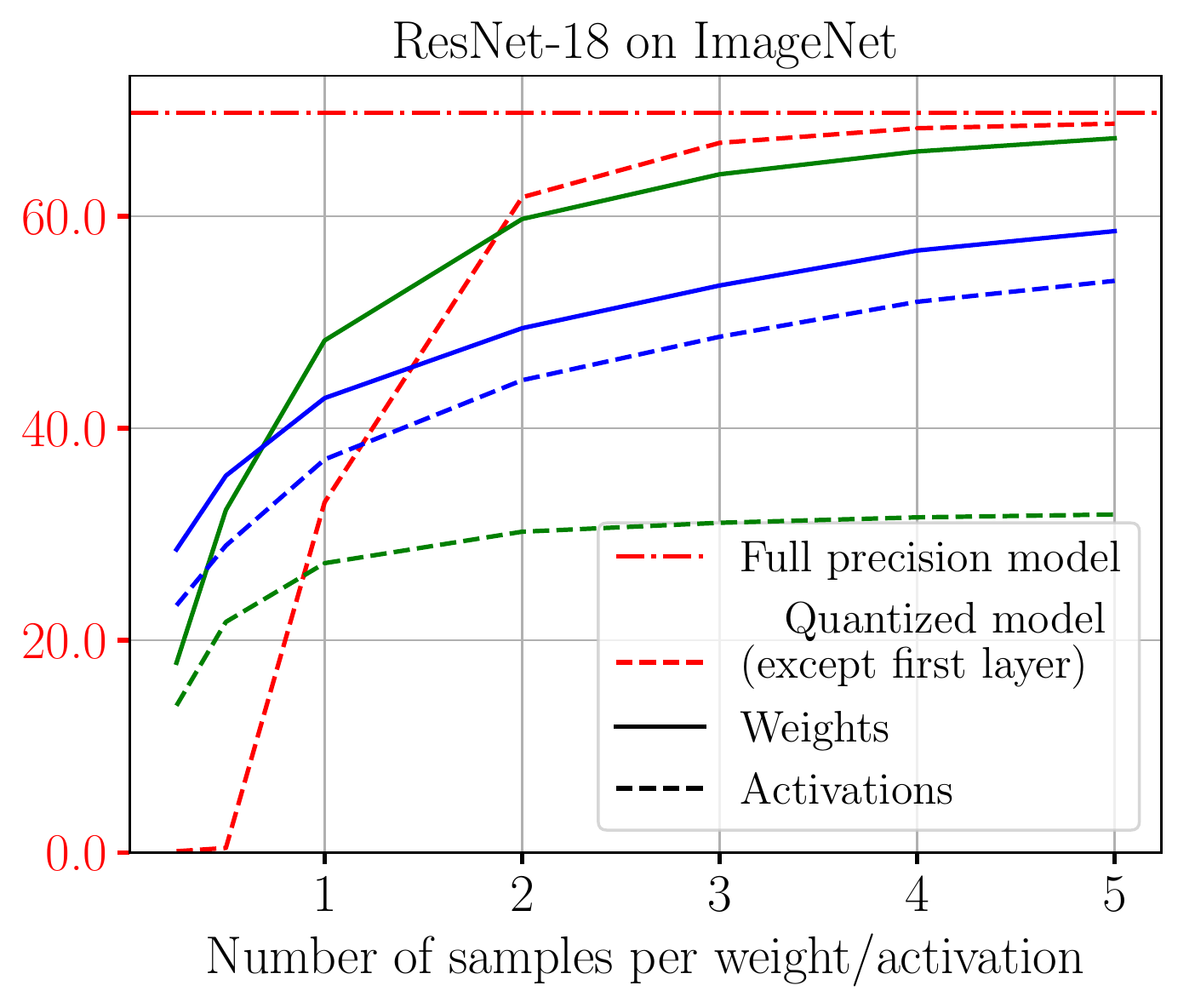}
        \end{subfigure}
        \begin{subfigure}{.305\linewidth}
            \includegraphics[width=\linewidth]{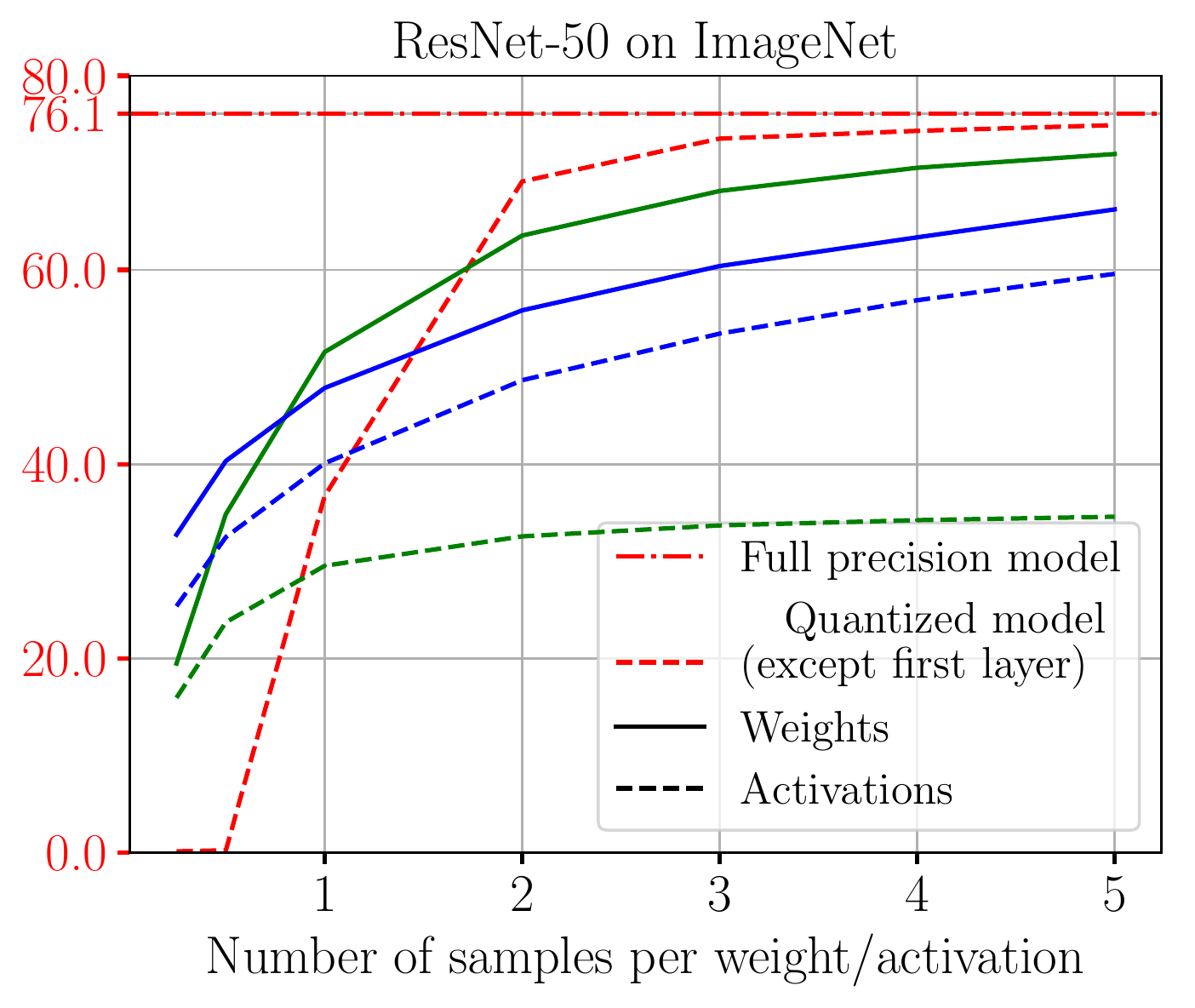}
        \end{subfigure}
        \begin{subfigure}{0.062\linewidth}
            \includegraphics[width=\linewidth]{Images/imagenet/weights/rightax}
        \end{subfigure}
    \caption{Results of quantizing both weights and activations on ImageNet using different sampling amounts. All quantized models reach close to full-precision accuracy at $K=3$.} 
        \label{fig:imagenet}
    \end{figure*}

     \Cref{fig:imagenet} shows the accuracy of the quantized model when using different sample sizes, \textit{i.e.,} $K \in \left[0.25, ..., 5.0\right]$. We observe that more sampling is required to achieve a close to full-precision model accuracy on ImageNet. On this dataset, sorting the CDF before sampling didn't result in any improvements, so reported results are without sorting.
    All the quantized models achieve close to full-precision accuracy, though more samples are required than for the previous datasets resulting in a higher required bit-width.
    
    \begin{table*}[ht]
        \caption{Accuracy results on ImageNet when quantizing weights, activations, or both. When quantizing weights only, accuracy drops less than 1\% in all tested models. Quantizing only the activations generally leads to a lower accuracy loss compared to quantizing weights. Quantizing both weights and activations leads to an additional accuracy loss of $0.6\%$ in the worst case, \textit{i.e.} ResNet-50.}
        %\vskip 0.15in
        \label{tab:imagenet}
        \begin{adjustbox}{width=1.0\textwidth}
        \centering
        \begin{small}
        \begin{sc}
        \begin{tabular}{lcccr}
            \toprule
            
            Method                          & AlexNet           & ResNet-18         & ResNet-50 \\
            \midrule
            Full Precision  (32w-32a)                 & 56.52             & 69.76             & 76.13 \\
            $\Delta$ MCQ (quantized w)     
                & -0.99 (8.00w-32a) / -0.68\footref{not_first_layer}  (8.00w-32a)       
                & -0.72 (8.00w-32a)/ -0.63\footref{not_first_layer} (8.00w-32a)        
                & -0.73 (8.28w-32a)/ -0.20\footref{not_first_layer} (8.28w-32a)\\
            
            $\Delta$ MCQ (quantized a)     
                & +0.02\footref{not_first_layer} (32w-8.36a)        
                & -0.58\footref{not_first_layer} (32w-7.36a)     
                & -0.76\footref{not_first_layer} (32w-7.45a)\\
            
            $\Delta$ MCQ (quantized w + a) 
                & -1.05 (7.88w-8.46a) / -0.75\footref{not_first_layer}  (8.00w-7.2a)  
                & -1.13 (8.00w-7.35a) /-1.03\footref{not_first_layer} (8.00w-7.36a) 
                & -1.64 (8.26w-7.43a) / -1.21\footref{not_first_layer} (8.28w-7.45a)\\
            
            \midrule
            
            $\Delta$ FGQ (2w-8a)  
                                        & -7.79\footref{first_layer_special} 
                                        & - 
                                        & -4.29 \\
            $\Delta$ TTQ (2w-32a) 
                                        & +0.3\footref{not_first_layer}$^,$\footref{not_last_layer} 
                                        & -3.0\footref{not_first_layer}$^,$\footref{not_last_layer}
                                        & - \\
            $\Delta$ TWNs (2w-32a) 
                                        & -2.7\footref{not_first_layer}$^,$\footref{not_last_layer} 
                                        & -4.3\footref{not_first_layer}$^,$\footref{not_last_layer} 
                                        &  - \\
            $\Delta$ BWN (1w-32a)  
                                        & +0.2
                                        & -8.5\footref{not_first_layer}$^,$\footref{not_last_layer} 
                                        & - \\
            $\Delta$ XNOR-Net (1w-1a) 
                                        & -12.4
                                        & -18.1\footref{not_first_layer}$^,$\footref{not_last_layer} 
                                        & - \\ 
            $\Delta$ DoReFa (1w-32a) 
                                        & -3.3\footref{not_first_layer}$^,$\footref{not_last_layer}  
                                        & - 
                                        & -\\
            $\Delta$ INQ (5w-32a) 
                                        & -0.15             & -0.71                 & -1.59\\
            $\Delta$ RQ (8w-8a) 
                                        & -                 & +0.43                 & -\\ 
            $\Delta$ LR-net (2w-32a) 
                                        & -                 & -6.07\footref{not_first_layer} & -\\ 
            \bottomrule
        \end{tabular}
        \end{sc}
        \end{small}
        \end{adjustbox}
        %\vskip -0.1in
        \label{tab:results_imagenet}
    \end{table*}
    
\subsection{Experiments on additional tasks}\label{sec:other_tasks}
    To assess the robustness of MCQ, we further evaluate MCQ on several models in natural language and speech processing. We evaluate language modeling on Wikitext-103 using a Transformer-based model~\cite{baevskiLM2018} and Wikitext-2 using a 2-layer LSTM~\cite{Zhao2019OCS}, speech recognition on VCTK using Deepspeech2~\cite{deepspeech2}, and machine translation on WMT-14 English-to-French using a Transformer~\cite{ott2018NMT}. Additional details are provided in the Appendix. Table~\ref{tab:results_nlp} shows the comparison to full-precision models for these various tasks.

    \begin{table*}[h!]
        \caption{Evaluation of MCQ on language modeling, speech recognition, and machine translation. 
        All quantized models reach close to full precision performance. Note that, as opposed to the image classification task, we did not study different sampling amounts nor the effect of quantization on specific network layers. A more in-depth analysis could then help to achieve close to full-precision accuracy at a lower bit-width on these models.
        }
        \begin{adjustbox}{width=1.0\textwidth}
        \centering
        \begin{small}
        \begin{sc}
        \begin{tabular}{llllcc}
        \toprule
        Task & Dataset & Model & Metric & Full Precision (32w-32a) & $\Delta$ MCQ (quantized w)\\
        \midrule
        Language Modeling   & WikiText-103  &   Transformer &   Perplexity $\downarrow$  &   18.7    &  +0.21 (8.21w-32a)\\
        Language Modeling   & WikiText-2    &   LSTM 2x650  &   Perplexity $\downarrow$  &   71.05   &   +0.51 (7.17w-32a)\\
        Speech Recognition  & VCTK          &   DeepSpeech2 &   CER $\downarrow$         &   7.00    &   +0.09 (7.26w-32a)    \\
        Machine Translation & WMT14 en-fr   &   Transformer &   BLEU $\uparrow$        &   40.83   &   -0.23 (7.71w-32a)    \\
        \bottomrule
        \end{tabular}
        \end{sc}
        \end{small}
        \end{adjustbox}
        \label{tab:results_nlp}
    \end{table*}

\section{Discussion and Future Work}\label{sec:discussion}
    
    The experimental results show the performance of MCQ on multiple models, datasets, and tasks, demonstrated by the minimal loss of accuracy compared to the full-precision counterparts. MCQ either outperforms or is competitive to other methods that require additional training of the quantized network. Moreover, the trade-off between accuracy, sparsity, and bit-width can be easily controlled by adjusting the number of samples. Note that the complexity of the resulting quantized network is proportional to the number of samples in both space and time.
    
    One limitation of MCQ, however, is that it often requires a higher number of bits to represent the quantized values. On the other hand, this sampling-based approach directly translates to a good approximation of the real full-precision values without any additional training. Recently \citet{Zhao2019OCS} proposed to outlier channel splitting, which is orthogonal work to MCQ and could be used to reduce the bit-width required for the highest hit counts.
    
    There are several paths that could be worth following for future investigations. 
    In the importance sampling stage, using more sophisticated metrics for importance ranking, \textit{e.g.} approximation of the Hessian by Taylor expansion could be beneficial ~\cite{molchanov2016pruning}. 
    Automatically selecting optimal sampling levels on each layer could lead to a lower cost since later layers seem to tolerate more sparsity and noise. 
    For efficient hardware implementation, it's important that the quantized network can be executed using integer operations only. Bias quantization and rescaling, activation rescaling to prevent overflow or underflow, and quantization of errors and gradients for efficient training leave room for future work.  
    
\clearpage
\section{Conclusion}
    
    In this work, we showed that Monte Carlo sampling is an effective technique to quickly and efficiently convert floating-point, full-precision models to integer, low bit-width models. Computational cost and sparsity can be traded for accuracy by adjusting the number of sampling accordingly.
    
    Our method is linear in both time and space in the number of weights and activations, and is shown to achieve similar results as the full-precision counterparts, for a variety of network architectures, datasets, and tasks. In addition, MCQ is very easy to use for quantizing and sparsifying any pre-trained model. It requires only a few additional lines of code and runs in a matter of seconds depending on the model size, and requires no additional training. The use of sparse, low-bitwidth integer weights and activations in the resulting quantized networks lends itself to efficient hardware implementations.

\clearpage
\bibliographystyle{plainnat}
\bibliography{neurips_2019_emc2_extended}

\clearpage
\appendix

\section{Algorithm}

An overview of the proposed method is given in Algorithm~\ref{alg:method}.

\begin{algorithm}
    \SetAlgoLined
    \KwIn{Pre-trained full-precision network}
    \KwOut{Quantized network with integer weights}
    \For{K=0 to L-1}{
    $unsorted_{idxs} \gets argsort(W_K)$\;
    $W_{sorted} \gets sort(W_K)$\;
    $W_{abs} \gets abs(W_{sorted})$\;
    \tcp{Create PDF}
    $W_{PDF} \gets \dfrac{W_{abs}}{\| W_K \|_1}$\;
    \tcp{Create CDF}
    $W_{CDF} \gets \sum_{i=1}^{|W_{PDF}|} W_{PDF_i}$\;
    $N \gets ceil(|W_K|*K)$\;
    $start_{idx} \gets 0$\;
    $\xi \gets random(0, 1)$ \;
    \tcp{Initialize discrete weights with zeros}
    $W'_K \gets [0] \times |W_K|$\;
    \tcp{Start subsampling}
    \For{i=0 to $N-1$}{
    $x_i \gets \dfrac{i + \xi}{N}$\;
    $hit_{idx} \gets argmax(W_{CDF}[start_{idx}:] \geq x_i) + start_{idx}$\;
    $start_{idx}\gets hit_{idx} $\;
    
    $unsorted_{idx} \gets unsorted_{idxs}[hit_{idx}]$
    
    \tcp{Update counter}
    \eIf{$W_K[unsorted_{idx}] > 0$}{
    $W'_K[unsorted_{idx}]\texttt{++}$\;% 
    }{
    $W'_K[unsorted_{idx}]\texttt{--}$\;% 
    }
    }
 
    \tcp{Update to integer weights}
    $W_K \gets W'_K$\;
    \tcp{Update layer's precision}
    $B_{W_K} \gets 1+floor(\log_2(max(abs(W'_K))))+1$ \;
    }
    \caption{Monte Carlo Quantization (MCQ) on network weights. $L$ represents the number of trainable layers, $K$ indicates the percentage of samples to be sampled per weight. The process is performed equivalently for quantizing activations at inference time. Our algorithm is linear in both time and space in the number of weights and activations.}
    \label{alg:method}
\end{algorithm}

\section{Avoiding Exploding Activations}\label{subsec:activation_explosion}
    
        When using integer weights, care has to be taken to avoid overflows in the activations. For that, activations can be scaled using a dynamically computed shifting factor as in \cite{integerDNNs}. With Monte Carlo sampling, since we know the expected value of the next-layer activations, we can scale accordingly.

        \begin{align}
            \mathbf{E}(I_{0,i}) &= \frac{N_{samples_I}}{N_I} &
            \mathbf{E}(W_{0,j}) &= \frac{N_{samples_{W_0}}}{N_I \cdot N_{L_1}}
        \end{align}
        %clude{B-content}
        
        With the activation equation presented in Section~\ref{sec:network_normalization} and $N_I$ connections from the input layer to every neuron in the second layer:
        
        \begin{equation}
            \begin{split}
                \mathbf{E}(|a_{l,j}|) = \sum_{i=0}^{N_I-1} \mathbf{E}(W_{0,j}) \cdot \mathbf{E}(I_{0,i}) %\\
                % \mathbf{E}(|a_{l,j}|) = N_I \cdot \frac{N_{samples_{W_0}} \cdot N_{samples_I}}{N_I \cdot N_{L_1} \cdot N_I}
            \end{split}
        \end{equation}
        
        With $N_{samples_{W_0}} = K_w \cdot (N_I \cdot N_{L_1})$ and $N_{samples_I} = K_a \cdot N_I$:
        
        \begin{equation}
            \begin{split}
                \mathbf{E}(|a_{l,j}|) = N_I \cdot \frac{K_w \cdot (N_I \cdot N_{L_1}) \cdot K_a \cdot N_I}{N_I \cdot N_{L_1} \cdot N_I} = N_I \cdot K_w \cdot K_a%\\
            \end{split}
        \end{equation}
        
        The activations of a neuron need to be scaled by its number of inputs (the receptive field $F_{in}$), multiplied with the number of samples per weight and the number of samples per activation. 
        This is also valid for neurons in convolutional layers, where the receptive field is 3D, \textit{e.g.} $3\times3\times128$.
        
        Moreover, care must be taken to scale biases correctly, by taking both the scaling of weights and activations into account:
        
        \begin{equation}
        bias_{scaled} = bias \cdot \frac{N_{samples}}{\|W_{orig}\|_1} \cdot \frac{1}{F_{in}}
        \end{equation}

\section{Full-Precision Models Training Details}

The architectures and training details of all tested models for CIFAR-10, SVHN, and ImageNet are presented in Sections~\ref{sup:cifar}, \ref{sup:svhn}, and \ref{sup:imagenet}, respectively. Details of the additional experiments presented in Section~\ref{sec:other_tasks} are shown in Sections~\ref{sup:VCTK}, \ref{sup:Wikitext}, and \ref{sup:NMT}.

\subsection{CIFAR-10}\label{sup:cifar}

We trained our full-precision baseline models on the CIFAR-10 dataset \cite{cifar}, consisting of 50000 training samples. We evaluated both our full-precision and quantized models similarly on the rest of the 10000 testing samples.
The full-precision VGG-7 ($2\times128C3-MP2-2\times256C3-MP2-2\times512C3-MP2-1024FC-Softmax$) and VGG-14 ($2\times64C3-MP2-2\times128C3-MP2-3\times256C3-MP2-3\times512C3-MP2-3\times512C3-MP2-1024FC-Softmax$) models were trained using the code at
\href{https://github.com/bearpaw/pytorch-classification}{https://github.com/bearpaw/pytorch-classification}.
Each was trained for 300 epochs with the Adam optimizer, with a learning rate starting at 0.1 and decreased by factor 10 at epochs 150 and 225,  batch size of 128, and weights decay of 0.0005.
The ResNet-20 model uses the standard configuration described \cite{residualnetworks}, with 64, 128 and 256 filters in the respective residual blocks. We used more filters to increase the number of available weights in the first block to sample from. This could be similarly performed by sampling more on this specific model to reduce the accuracy loss. The ResNet-20 model is trained using the same hyperparameter settings as the VGG models. 

\subsection{SVHN}\label{sup:svhn}

We trained our full-precision baseline models on the Street View House Numbers (SVHN) dataset \cite{svhn}, consising of 73257 training samples. We evaluated both our full-precision and quantized models similarly using the 26032 testing samples provided in this dataset.
The full-precision VGG-7* model ($2\times64C3-MP2-2\times128C3-MP2-2\times256C3-MP2-1024FC-Softmax$) was trained for 164 epochs, using the Adam optimizer with learning rate starting at 0.001 and divided by 10 at epochs 80 and 120, weight decay 0.001, and batch size 200. Models A ($48C3-MP2-2\times64C3-MP2-3\times128C3-MP2-512C3-Softmax$), B, C, and D were trained using the code at \href{https://github.com/aaron-xichen/pytorch-playground}{https://github.com/aaron-xichen/pytorch-playground}
and the same hyperparameter settings as VGG-7* but trained for 200 epochs.

\subsection{ImageNet}\label{sup:imagenet}
We evaluated both our full-precision and quantized models similarly on the validation set of the ILSVRC12 classification dataset \cite{imagenet}, consisting of 50K validation images. The full-precision pre-trained models are taken from Pytorch's model zoo \href{https://pytorch.org/docs/stable/torchvision/models.html}{https://pytorch.org/docs/stable/torchvision/models.html} \citep{pytorch}. 

\subsection{VCTK}\label{sup:VCTK}
CSTR's VCTK Corpus (Centre for Speech Technology Voice Cloning Toolkit) includes speech data uttered by 109 native speakers of English with various accents, where each speaker reads out about 400 sentences, most of which were selected from a newspaper.
The evaluated model uses 2 convolutional layers and 5 GRU layers of 768 hidden units, using code from \href{https://github.com/SeanNaren/deepspeech.pytorch}{https://github.com/SeanNaren/deepspeech.pytorch} \citep{veaux2017cstr}.

\subsection{Wikitext}\label{sup:Wikitext}
The WikiText language modeling dataset is a collection of over 100 million tokens extracted from the set of verified Good and Featured articles on Wikipedia. Compared to the preprocessed version of Penn Treebank (PTB), WikiText-2 is over 2 times larger and WikiText-103 is over 110 times larger. The WikiText dataset also features a far larger vocabulary and retains the original case, punctuation and numbers - all of which are removed in PTB. As it is composed of full articles, the dataset is well suited for models that can take advantage of long term dependencies.
The WikiText-2 model was a 2-layer LSTM with 650 hidden neurons, and an embedding size of 400. It was trained using the setup and code at \href{https://github.com/salesforce/awd-lstm-lm}{https://github.com/salesforce/awd-lstm-lm} \citep{merityregularizing2017}.
The WikiText-102 model was a pretrained model available at \href{https://github.com/pytorch/fairseq/tree/master/examples/language_model}{https://github.com/pytorch/fairseq/tree/master/examples/language\_model}, along with evaluation code \citep{baevskiLM2018}.

\subsection{NMT}\label{sup:NMT}
The dataset is WMT’14 English-French, cmobining data from several other corpuses, amongst others the Europarl corpus, the News Commentary corpus, and the Common Crawl corpus \citep{machacek2014results}. 
The model was a pretrained model available at \href{https://github.com/pytorch/fairseq/tree/master/examples/scaling_nmt}{https://github.com/pytorch/fairseq/tree/master/examples/scaling\_nmt}, along with evaluation code \citep{ott2018NMT}.

\section{Quantizing Weights Only}
\begin{figure*}[ht] 
    \begin{subfigure}{.305\linewidth}
    	\includegraphics[width=\linewidth]{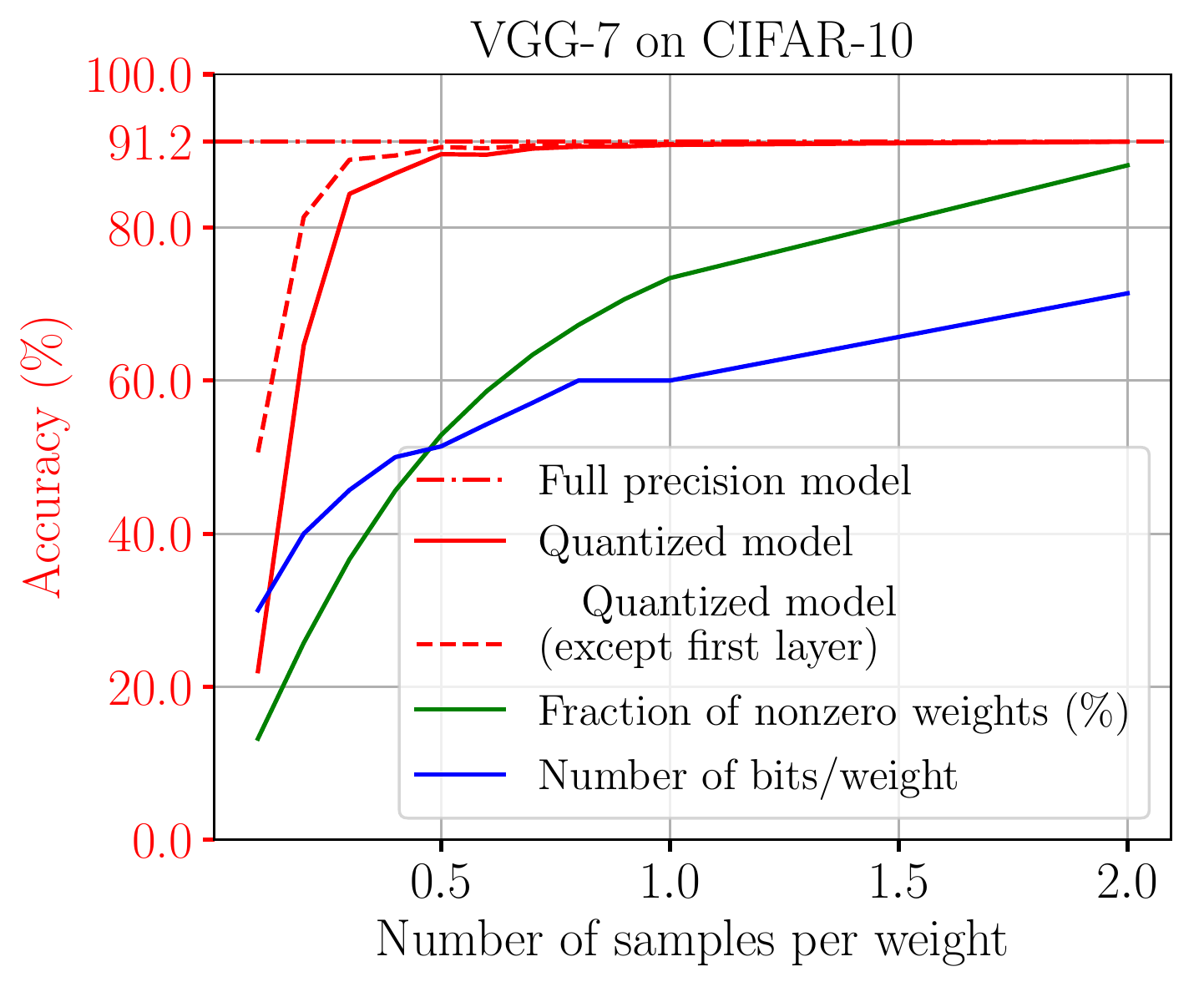}
    \end{subfigure}
    \begin{subfigure}{.305\linewidth}
    	\includegraphics[width=\linewidth]{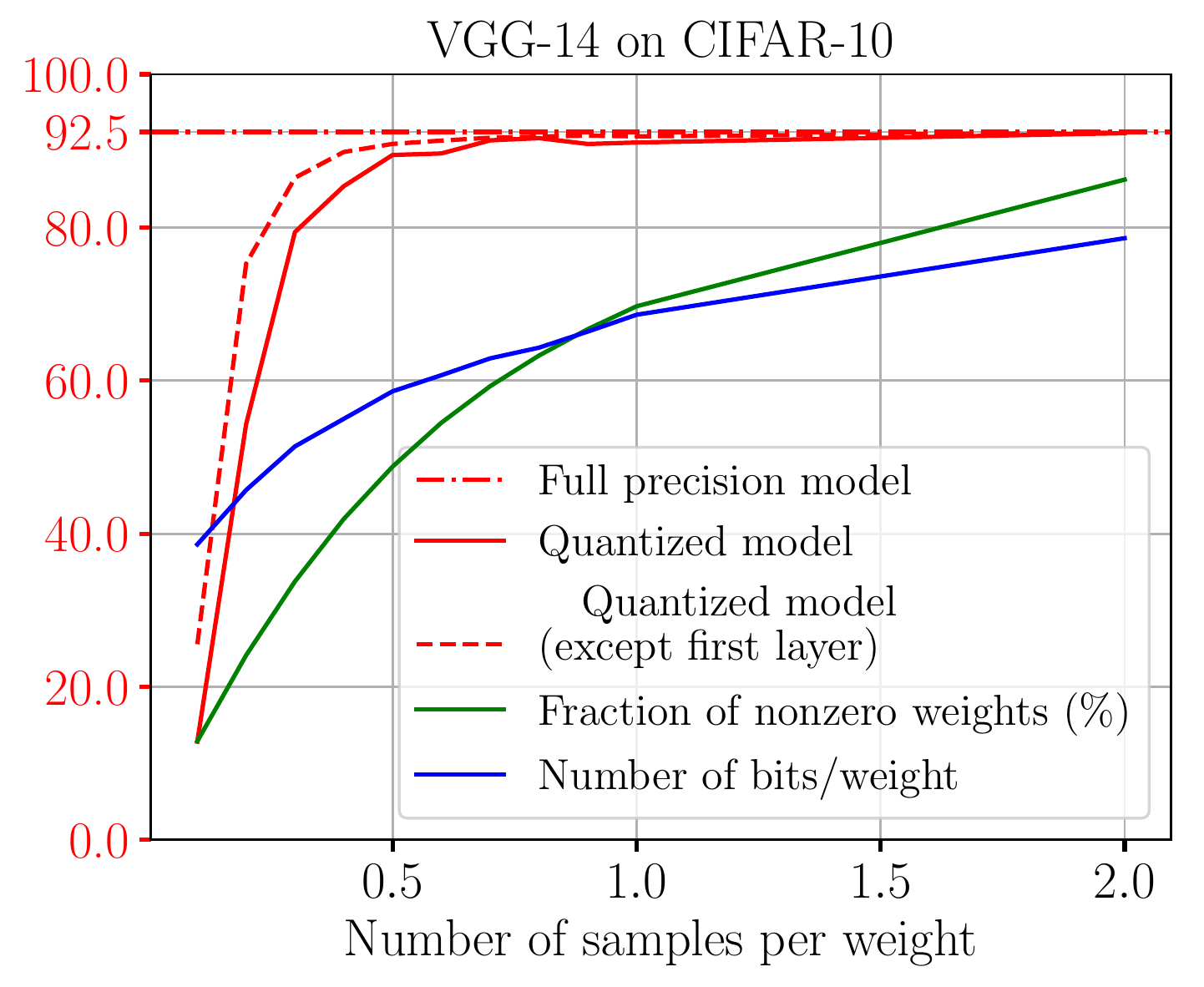}
    \end{subfigure} 
    \begin{subfigure}{.305\linewidth}
    	\includegraphics[width=\linewidth]{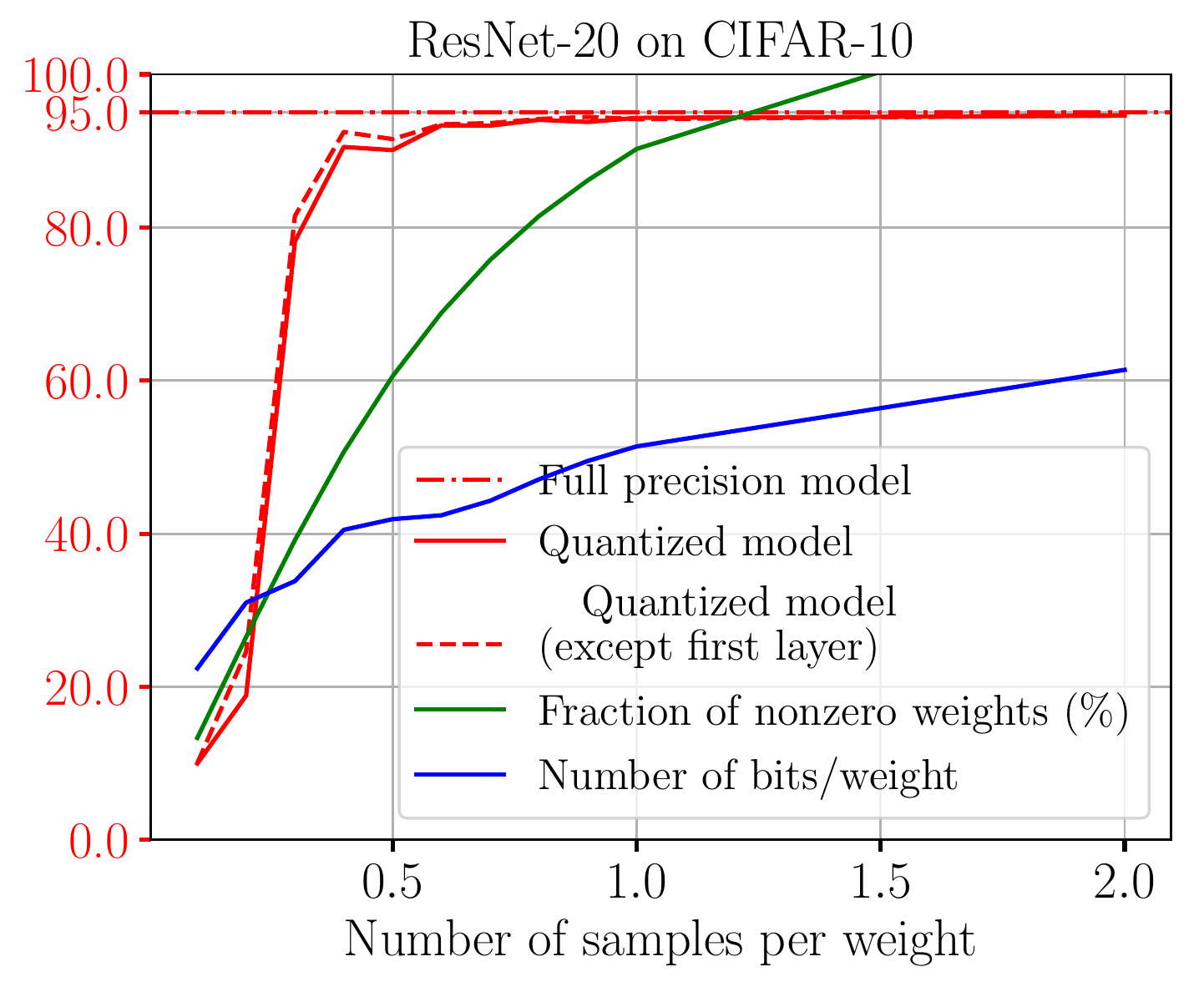}
    \end{subfigure}
    \begin{subfigure}{.062\linewidth}
    	\includegraphics[width=\linewidth]{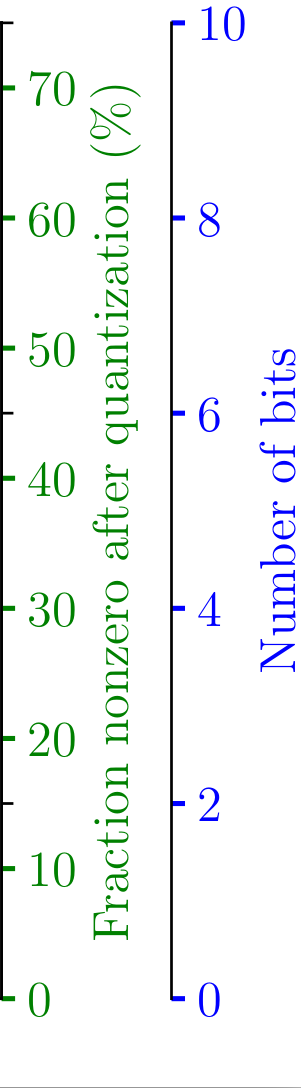}
    \end{subfigure}
    \caption{Quantized weights on CIFAR-10} 
    \label{sup:cifar_weights}
\end{figure*}

\begin{figure*}[ht] 
    \begin{subfigure}{.305\linewidth}
        \includegraphics[width=\linewidth]{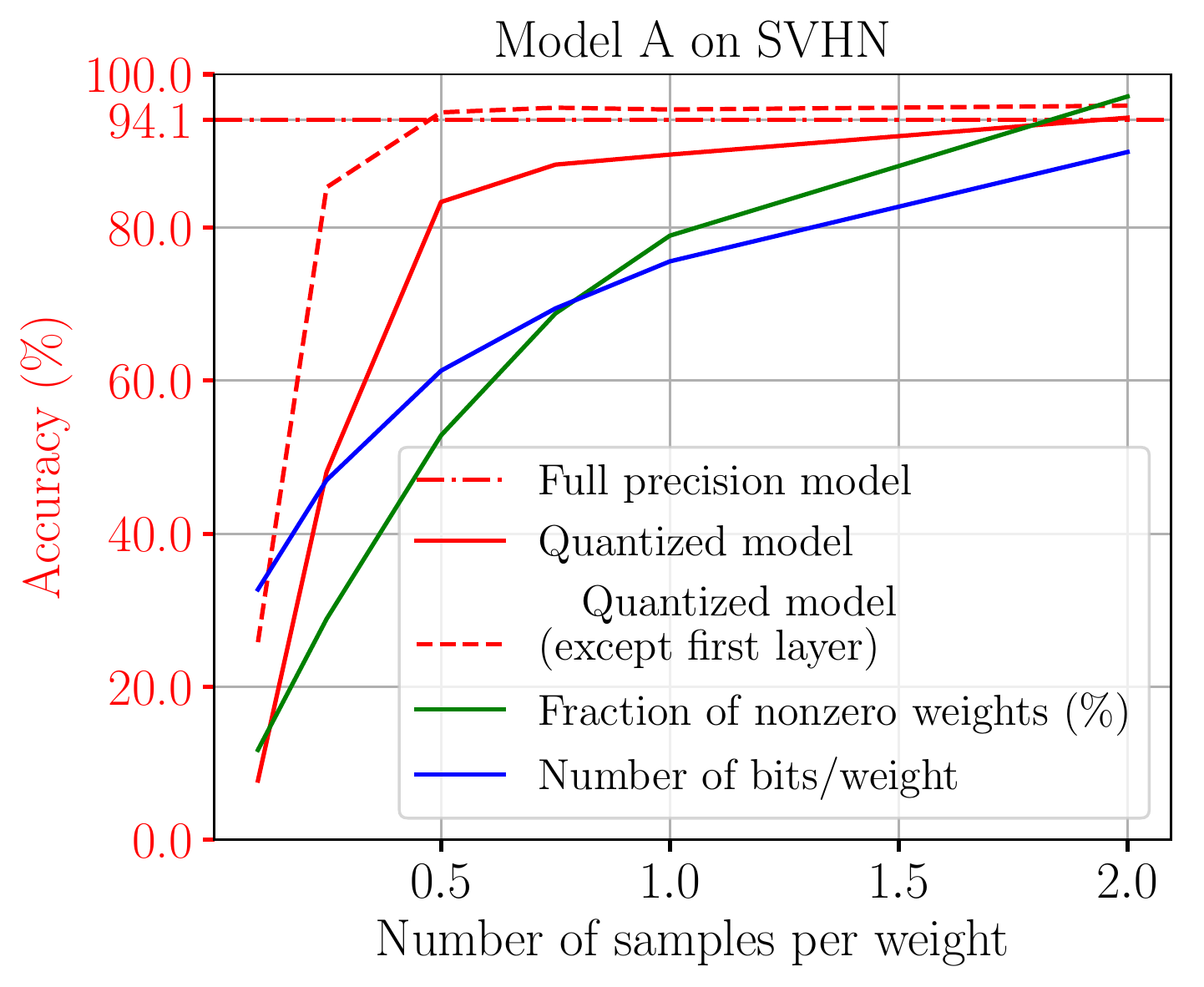}
    \end{subfigure}
   \begin{subfigure}{.305\linewidth}
        \includegraphics[width=\linewidth]{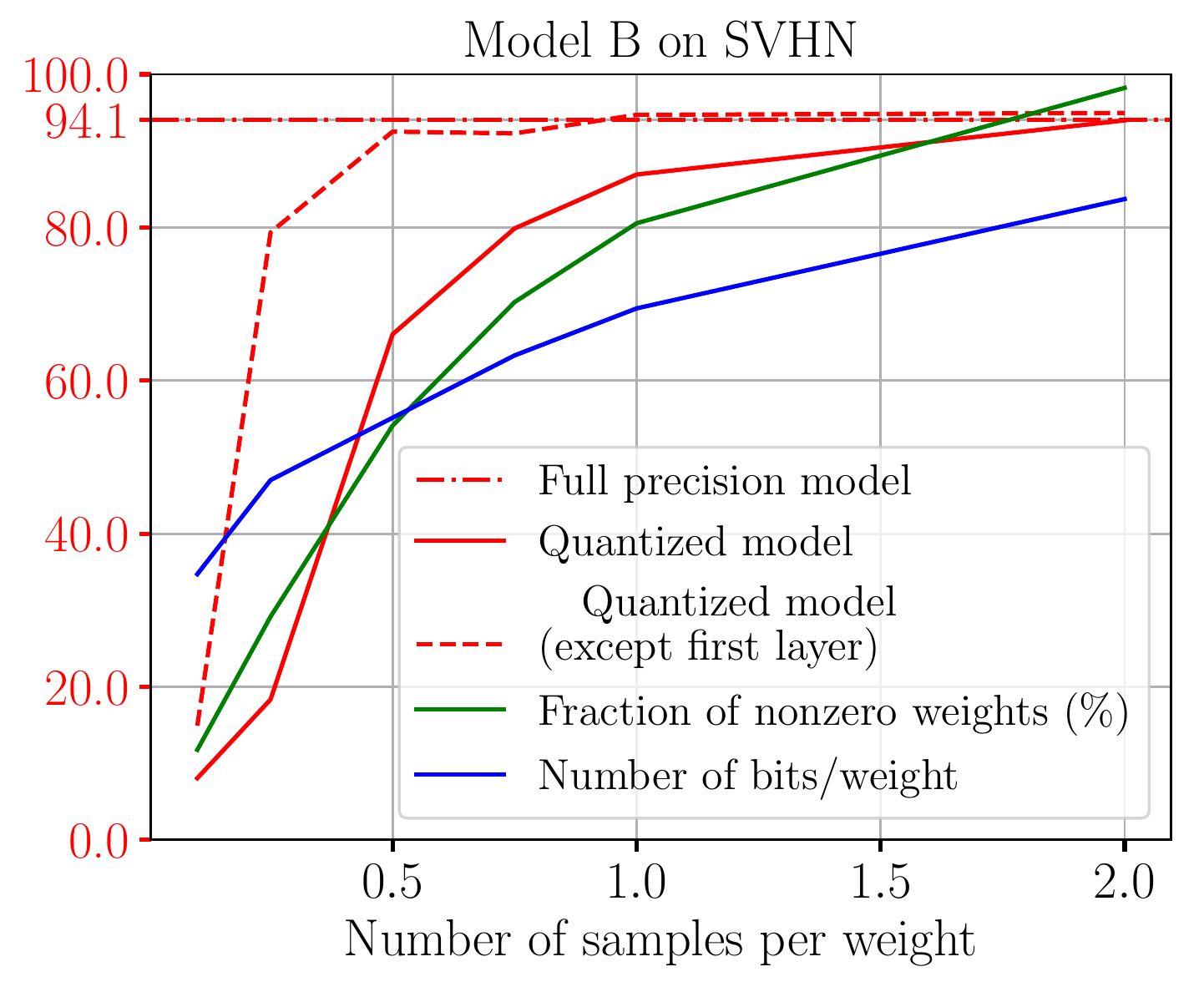}
    \end{subfigure} 
    \begin{subfigure}{.305\linewidth}
        \includegraphics[width=\linewidth]{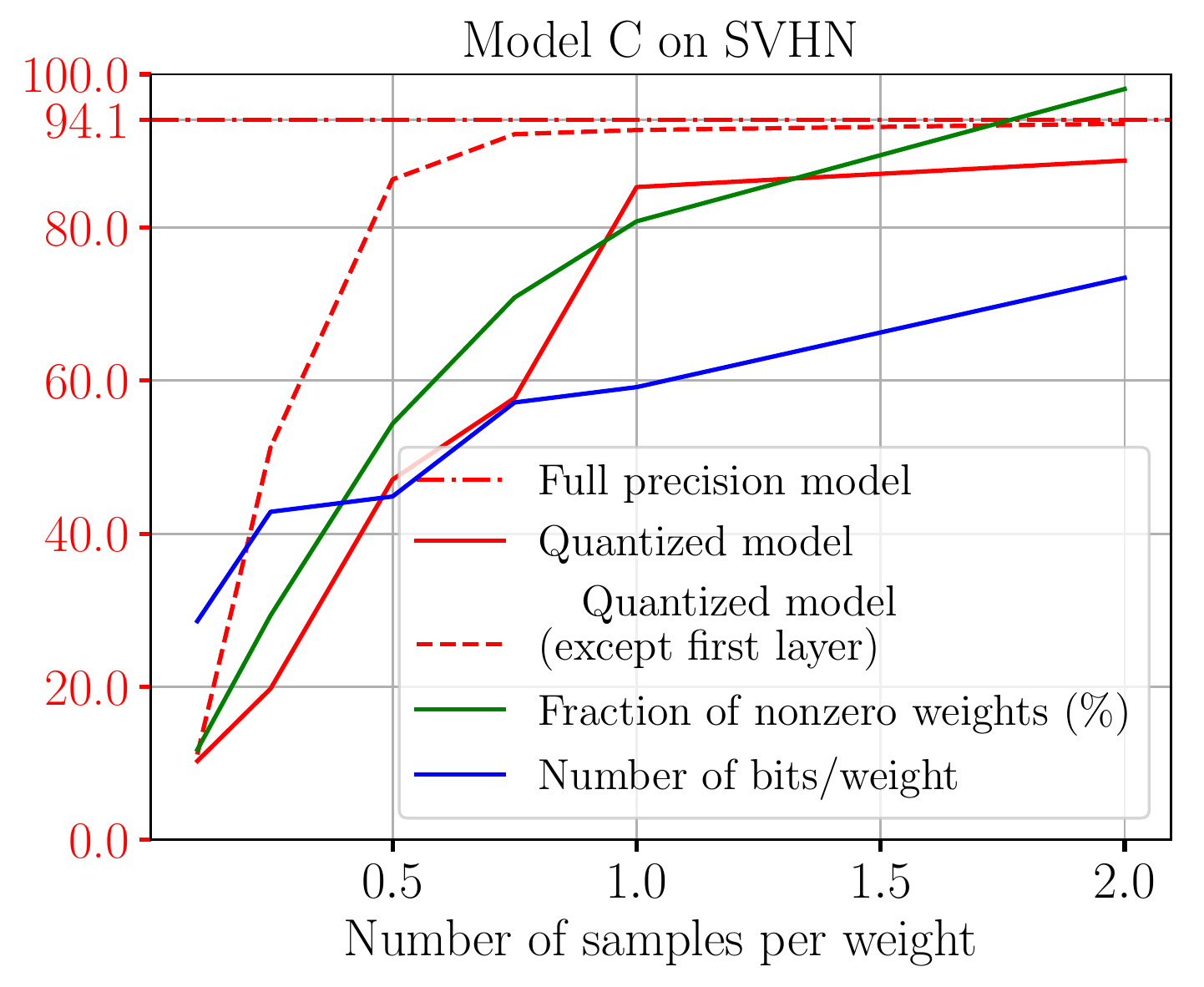}
    \end{subfigure} 
    \begin{subfigure}{.053\linewidth}
        \includegraphics[width=\linewidth]{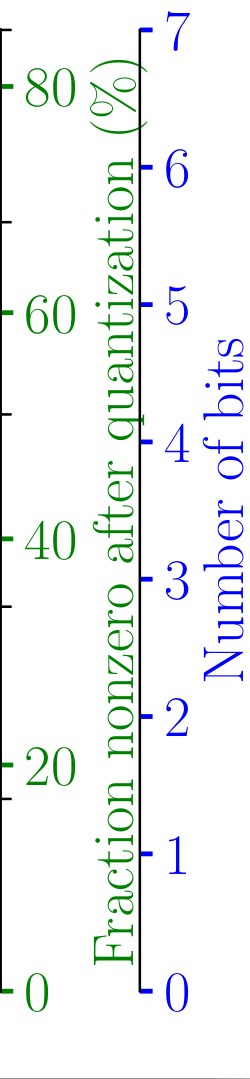}
    \end{subfigure} 

    \begin{center}
    \begin{subfigure}{.36\linewidth}
        \includegraphics[width=\linewidth]{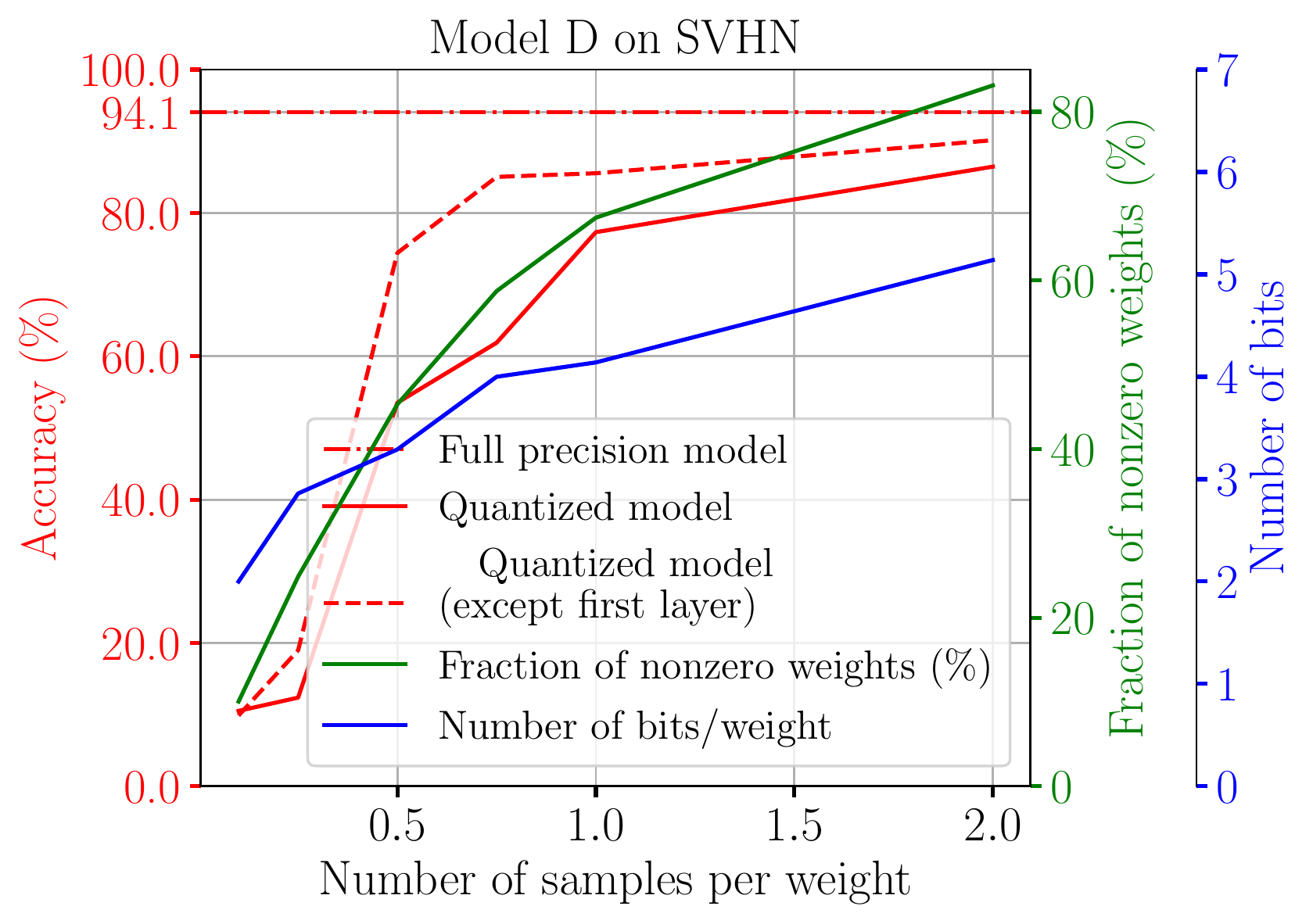}
    \end{subfigure}
    \begin{subfigure}{.36\linewidth}
        \includegraphics[width=\linewidth]{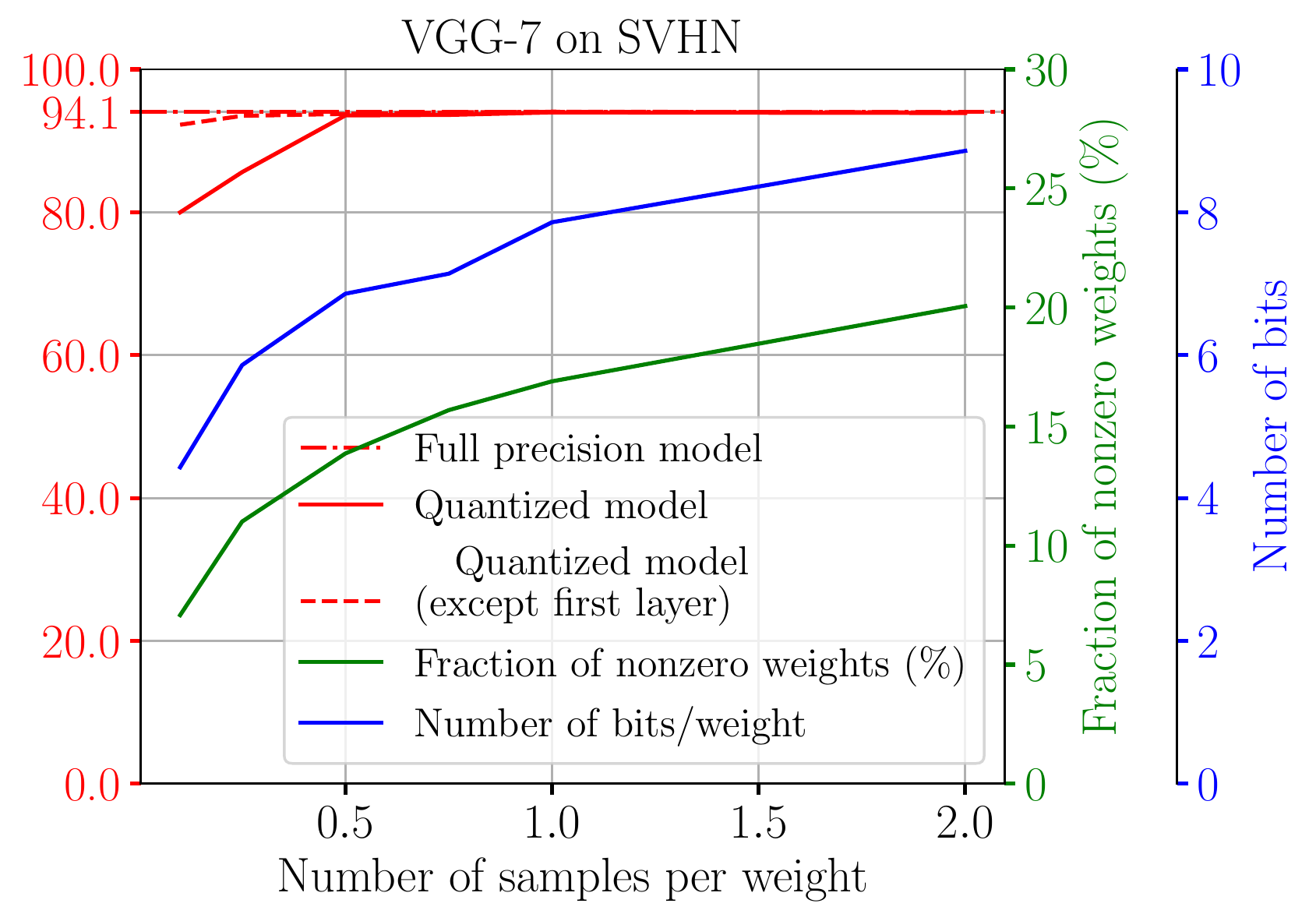}
    \end{subfigure}
    \end{center}
    \caption{Quantized weights on SVHN}
    \label{sup:svhn_weights}
\end{figure*}

\begin{figure*}[ht] 
    \begin{subfigure}{.305\linewidth}
	 	\includegraphics[width=\linewidth]{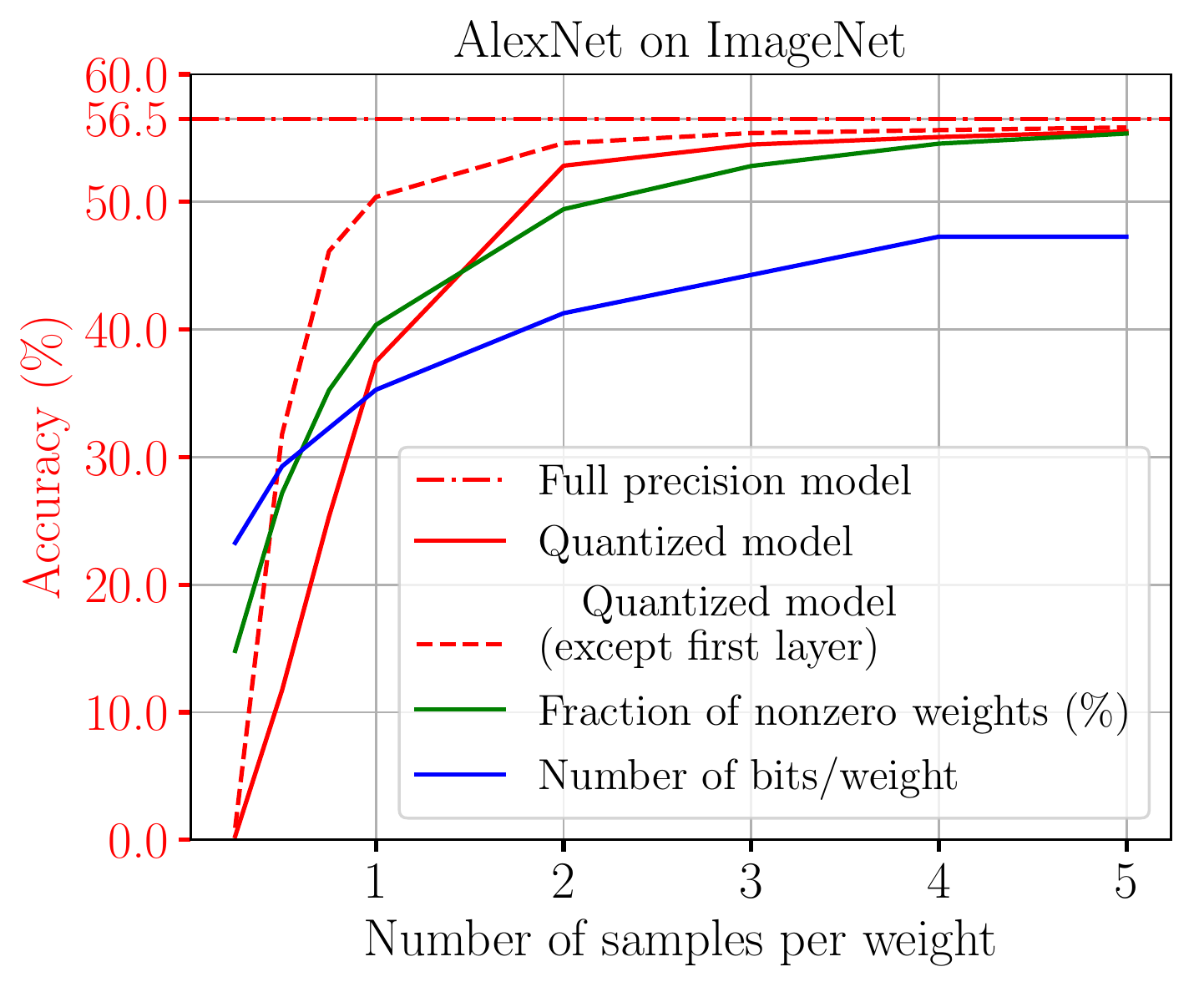}
	 \end{subfigure}
	 \begin{subfigure}{.305\linewidth}
	 	\includegraphics[width=\linewidth]{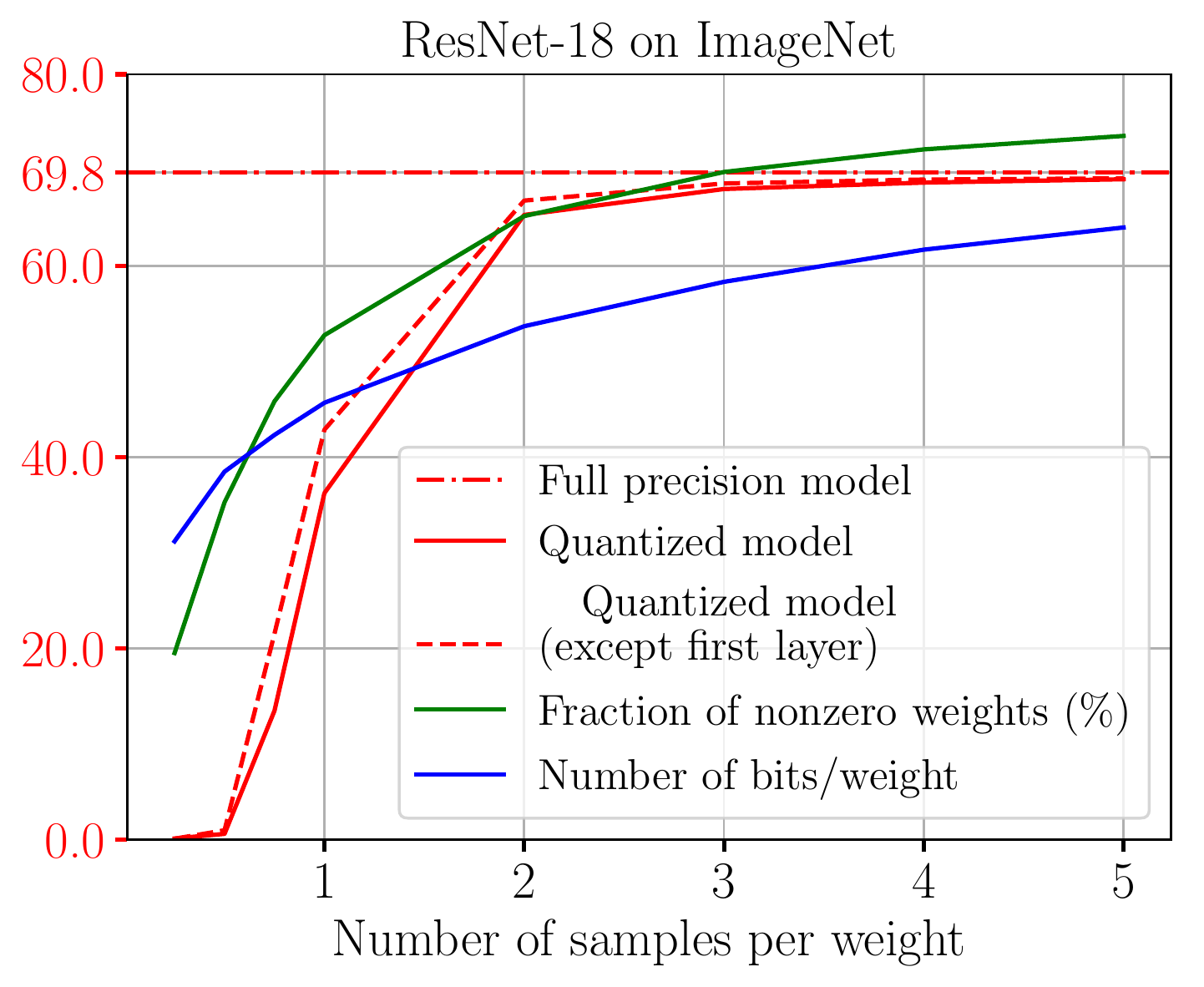}
	 \end{subfigure} 
	 \begin{subfigure}{.305\linewidth}
	 	\includegraphics[width=\linewidth]{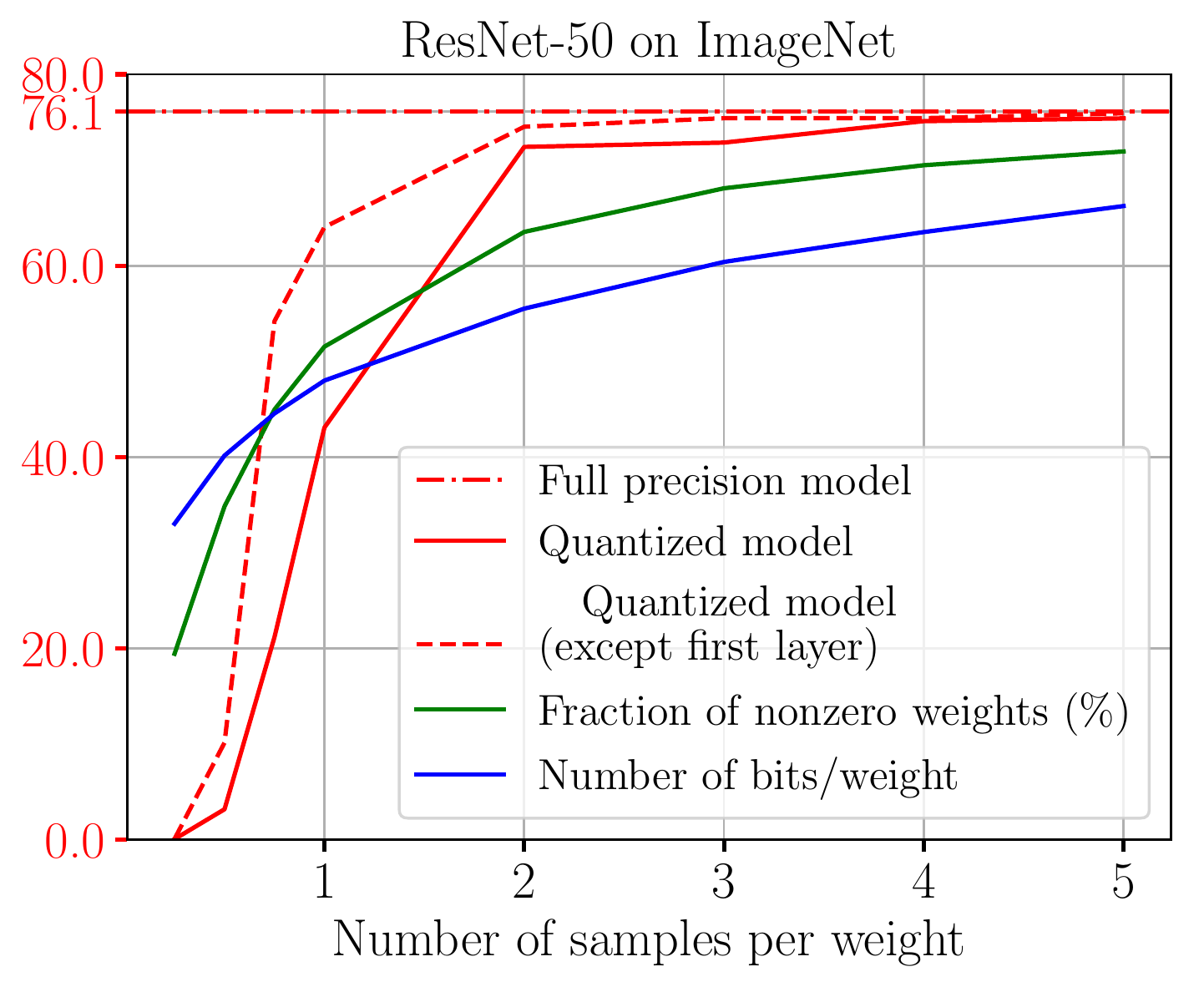}
	 \end{subfigure}
    \begin{subfigure}{.06\linewidth}
    	\includegraphics[width=\linewidth]{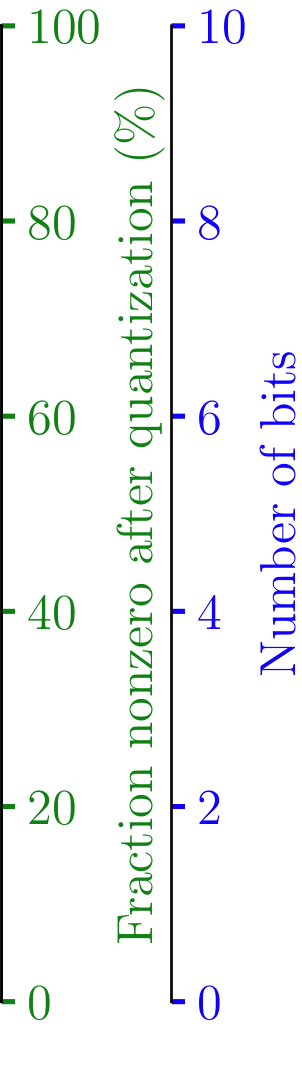}    
	\end{subfigure}
    \caption{Quantized weights on ImageNet} 
    \label{sup:imagenet_weights}
\end{figure*}

Figures \ref{sup:cifar_weights}, \ref{sup:svhn_weights}, and \ref{sup:imagenet_weights} show the effects of varying the amounts of sampling when quantizing only the weights.

\section{Quantizing Activations Only}

\begin{figure*}[ht] 
    \begin{subfigure}{.305\linewidth}
    	\includegraphics[width=\linewidth]{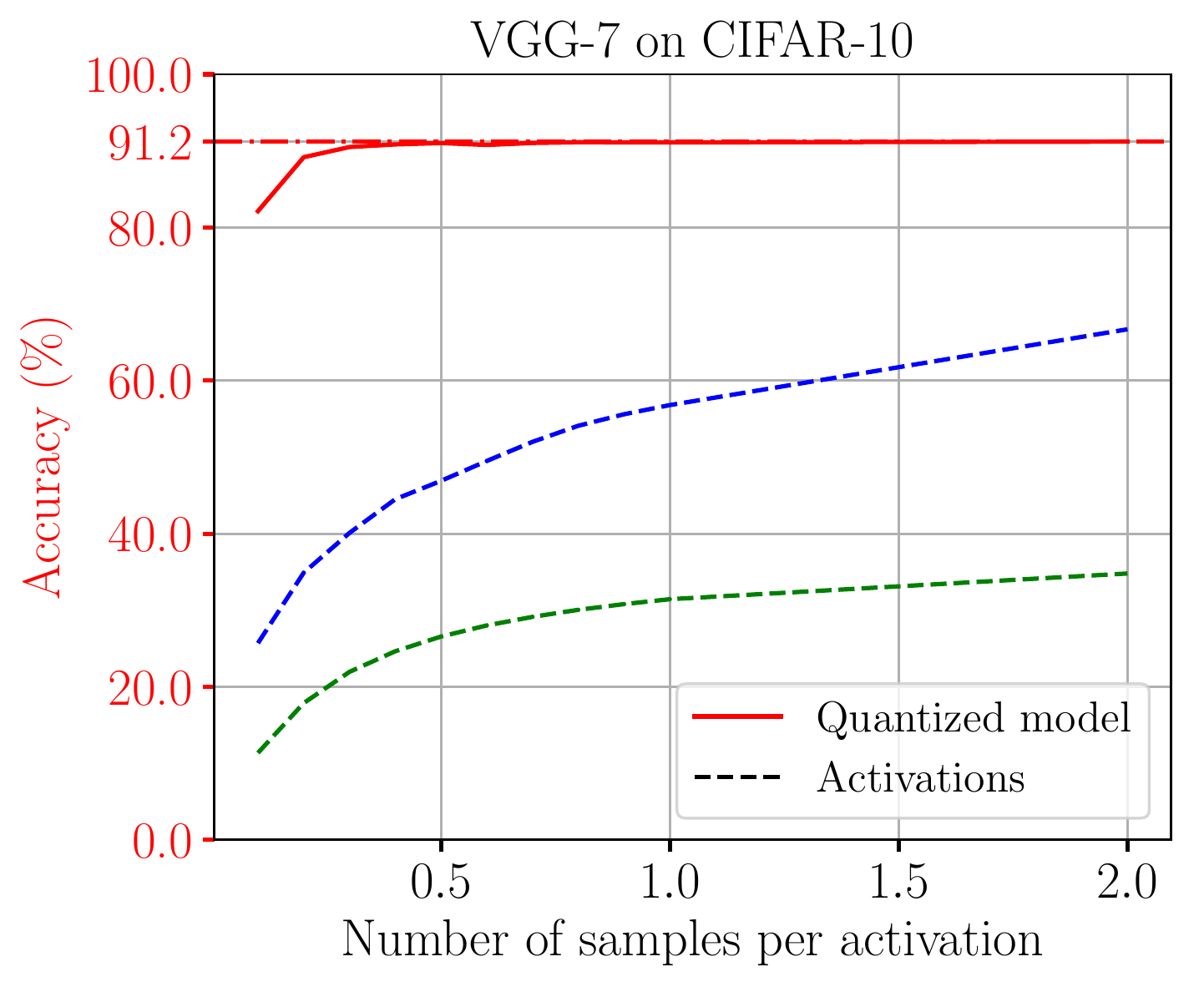}
    \end{subfigure}
    \begin{subfigure}{.305\linewidth}
    	\includegraphics[width=\linewidth]{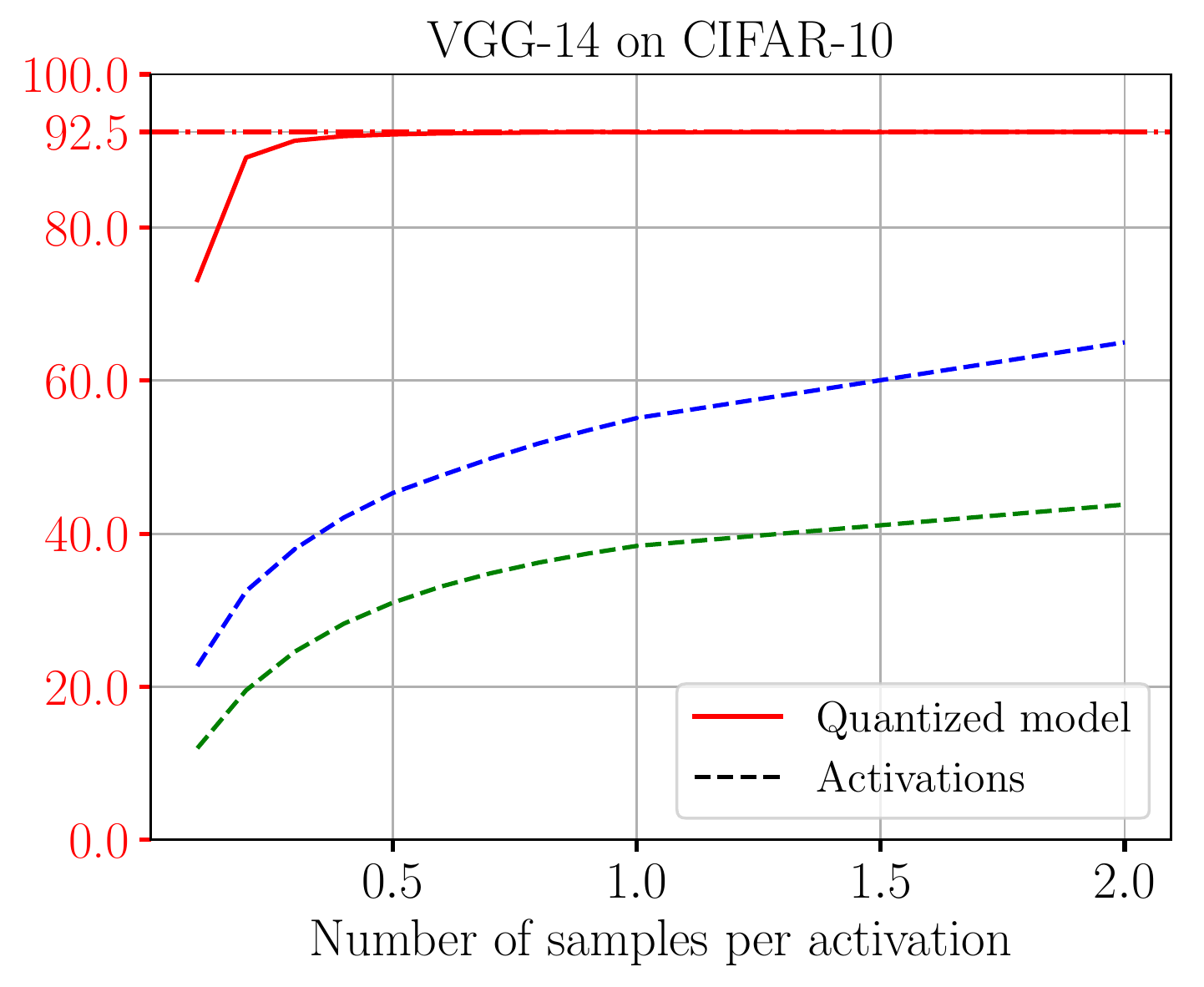}
    \end{subfigure} 
    \begin{subfigure}{.305\linewidth}
    	\includegraphics[width=\linewidth]{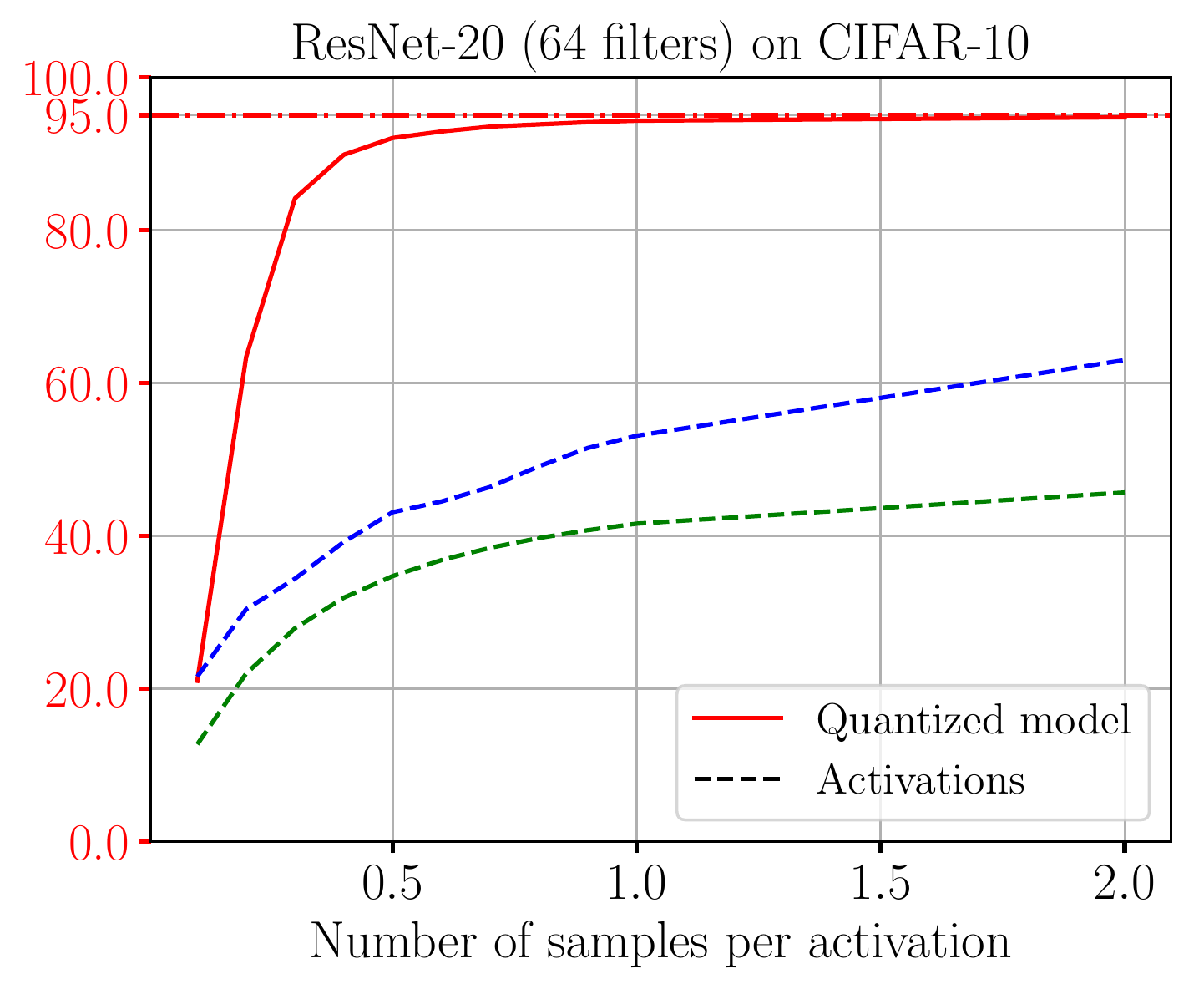}
    \end{subfigure}
    \begin{subfigure}{.062\linewidth}
    	\includegraphics[width=\linewidth]{Images/cifar10/weights/rightax.png}
    \end{subfigure}
    \caption{Quantized activations on CIFAR-10} 
    \label{sup:cifar_activations}
\end{figure*}

\begin{figure*}[ht] 
    
    \begin{subfigure}{.305\linewidth}
        \includegraphics[width=\linewidth]{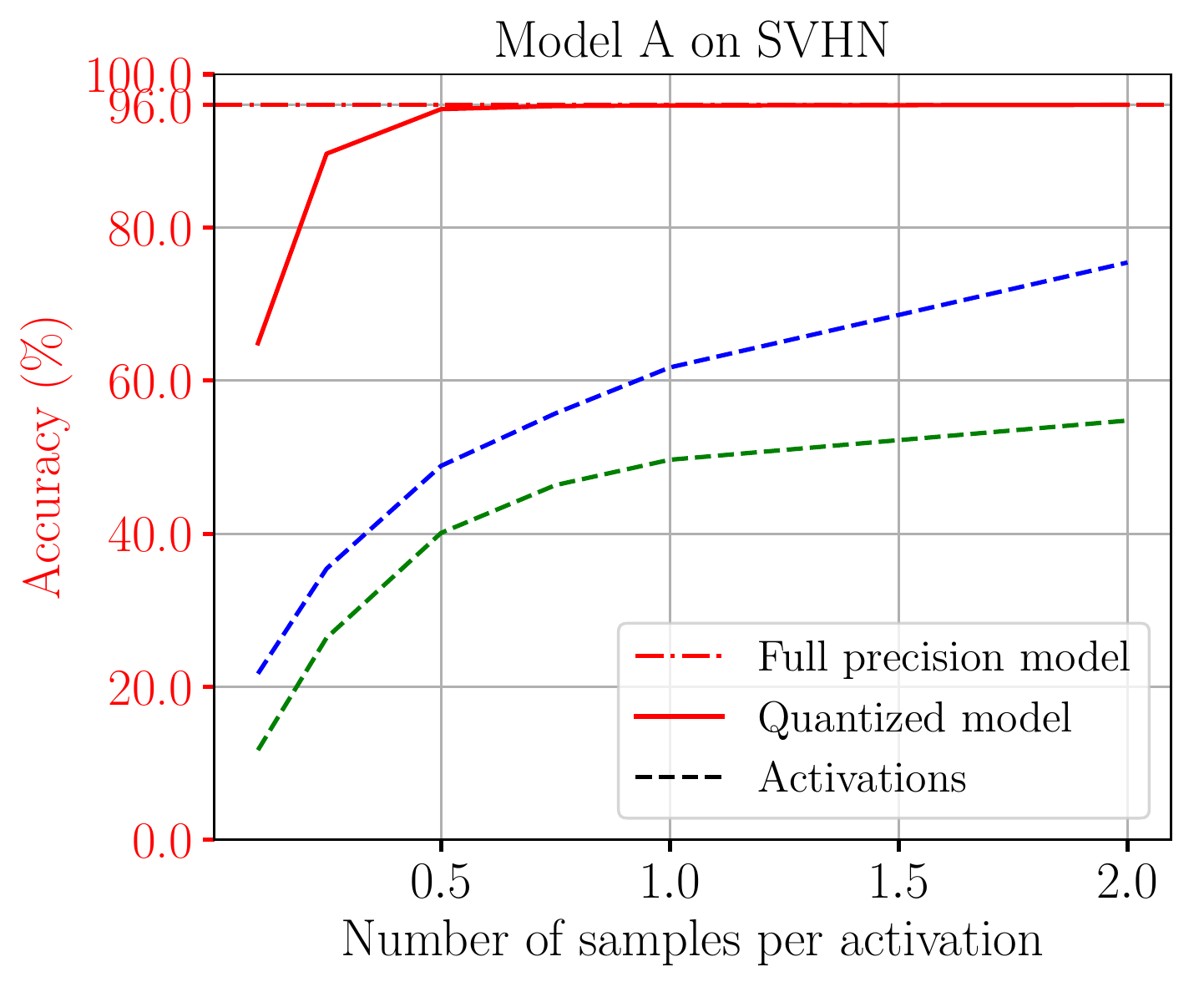}
    \end{subfigure}
   \begin{subfigure}{.305\linewidth}
        \includegraphics[width=\linewidth]{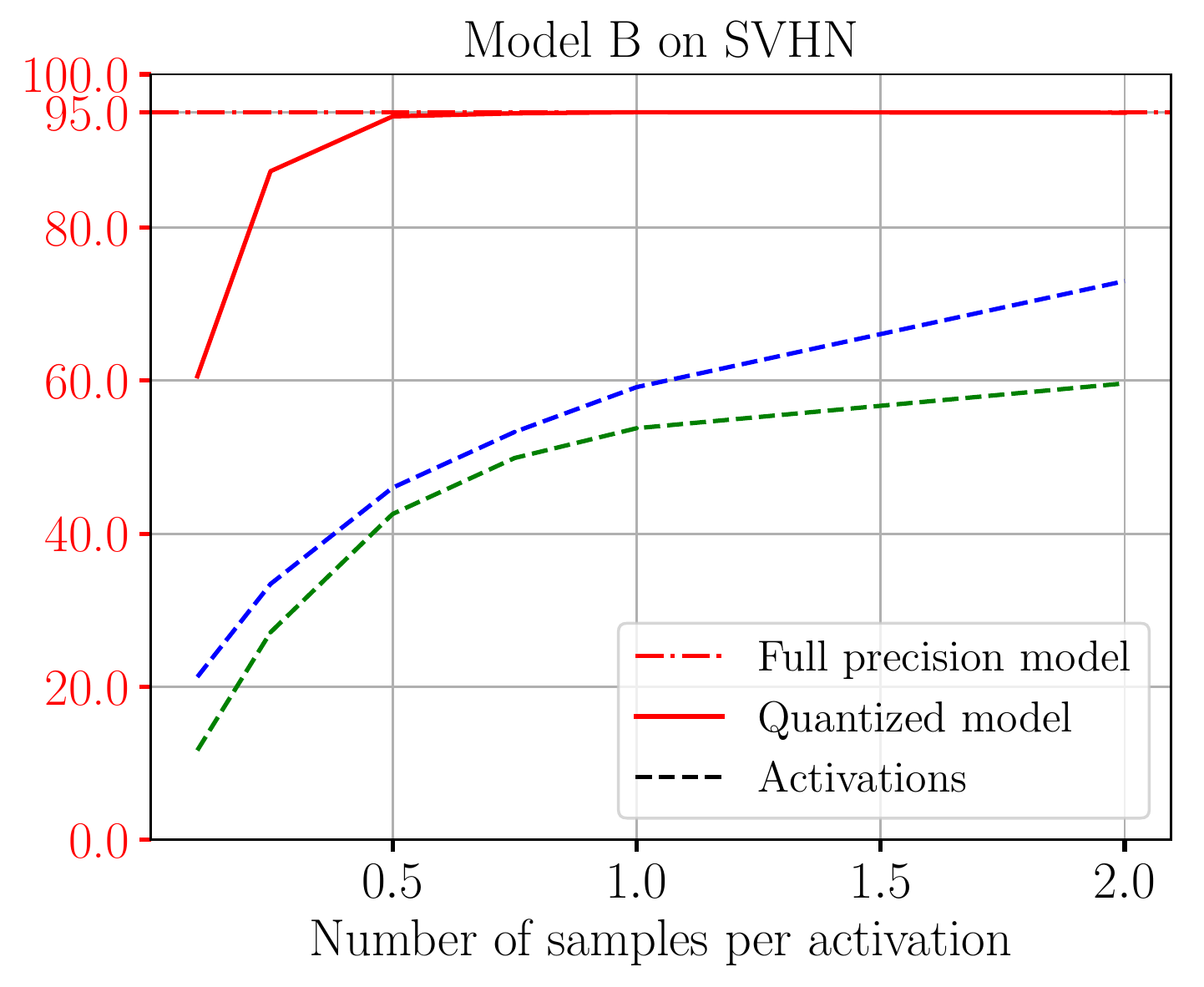}
    \end{subfigure} 
    \begin{subfigure}{.31\linewidth}
        \includegraphics[width=\linewidth]{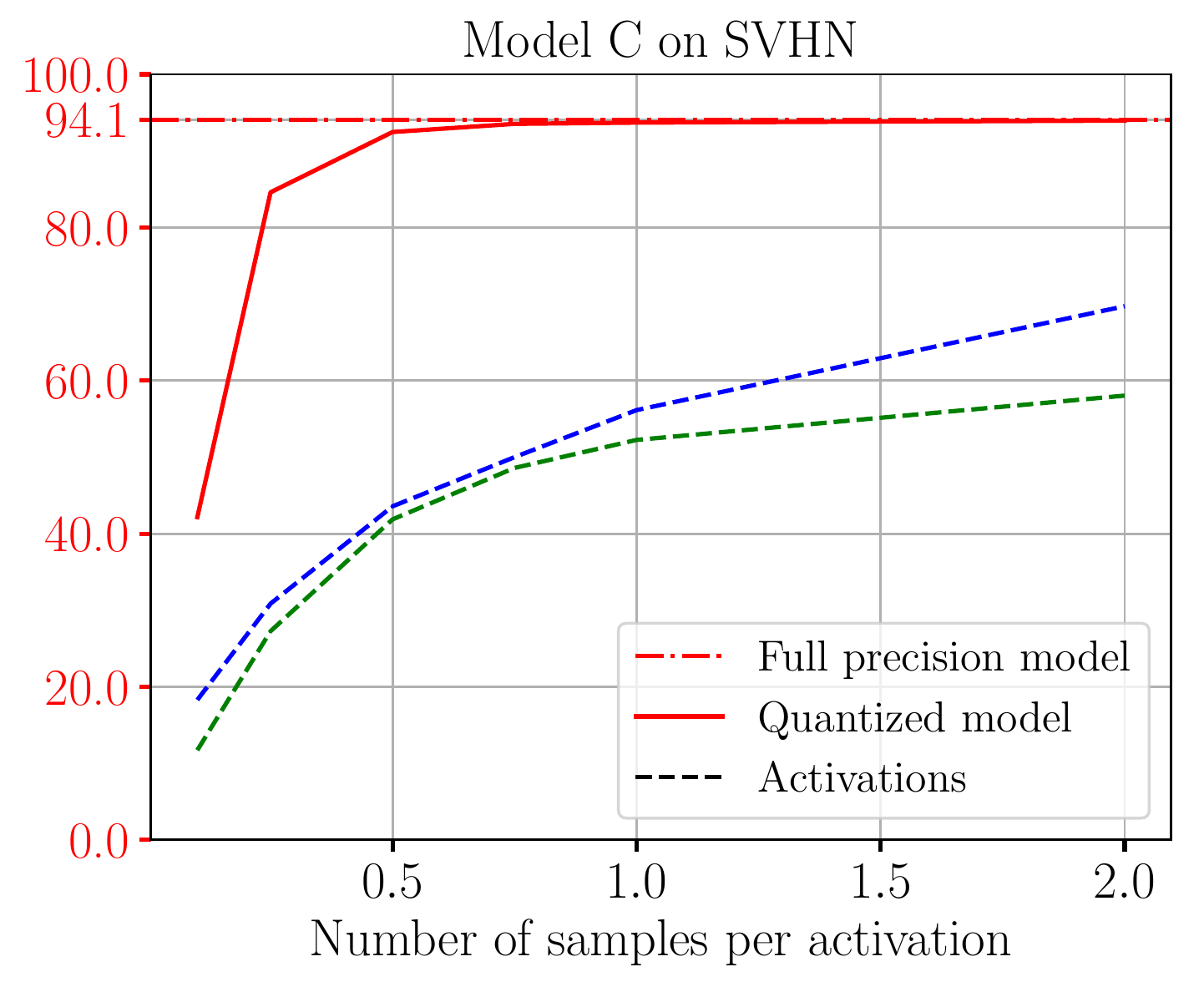}
    \end{subfigure} 
    \begin{subfigure}{.053\linewidth}
    \includegraphics[width=\linewidth]{Images/svhn/weights+activations/rightax.png}
    \end{subfigure}
    \begin{center} 
        \begin{subfigure}{.36\linewidth}
            \includegraphics[width=\linewidth]{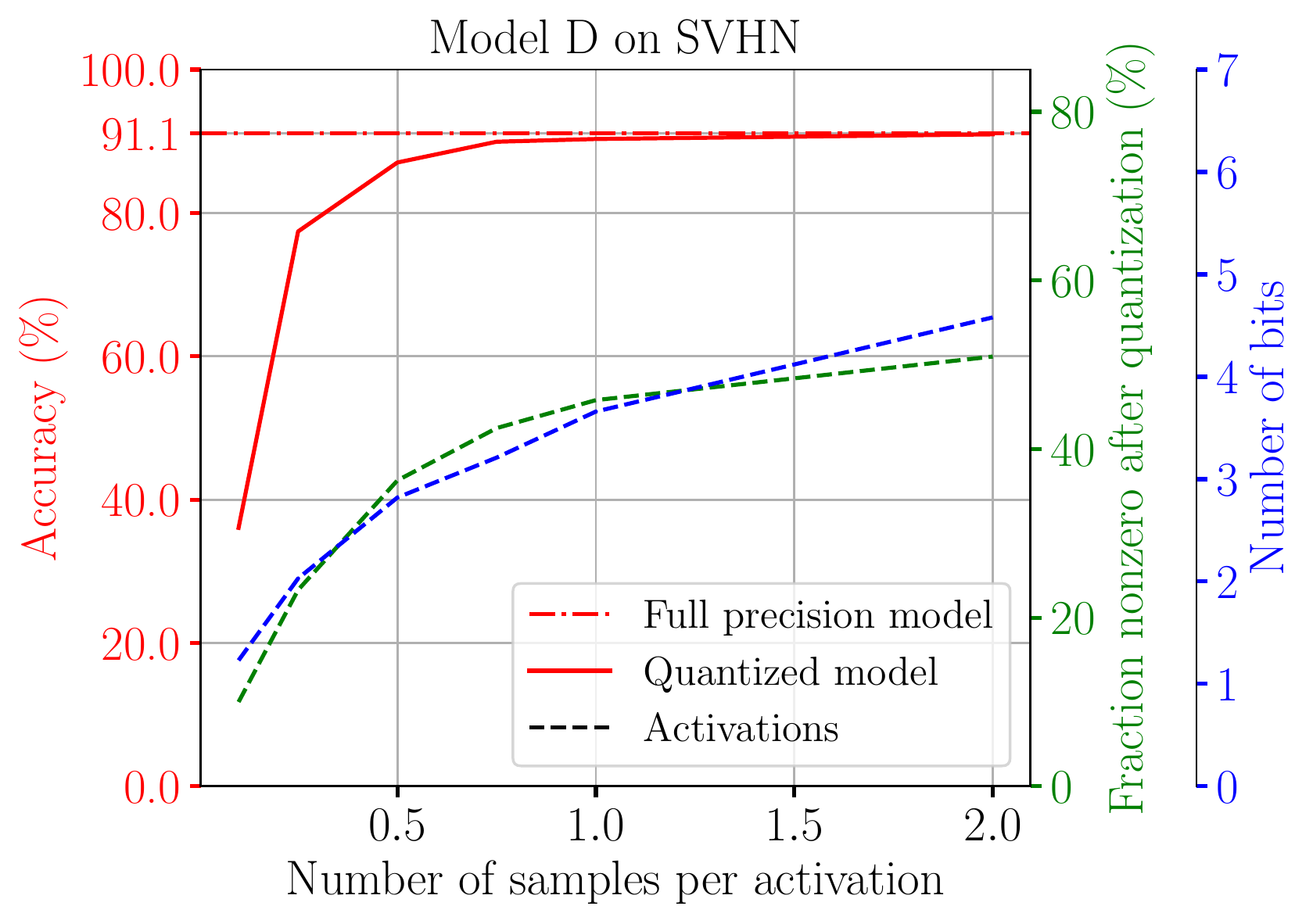}
        \end{subfigure}
        \begin{subfigure}{.36\linewidth}
            \includegraphics[width=\linewidth]{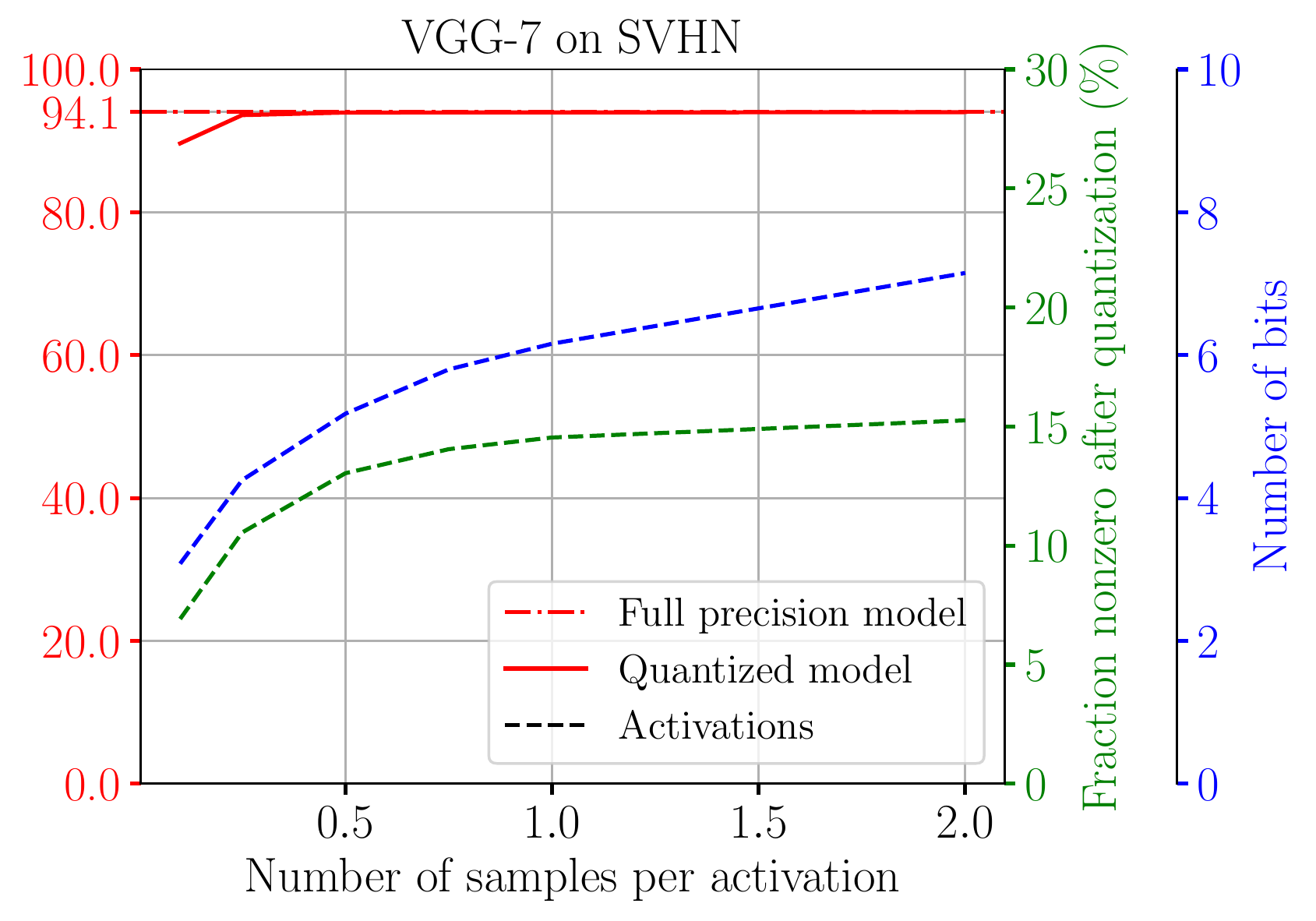}
        \end{subfigure}
    \end{center}
    \caption{Quantized activations on SVHN} 
    \label{sup:svhn_activations}
\end{figure*}

\begin{figure*}[ht] 
    \begin{subfigure}{.305\linewidth}
    	\includegraphics[width=\linewidth]{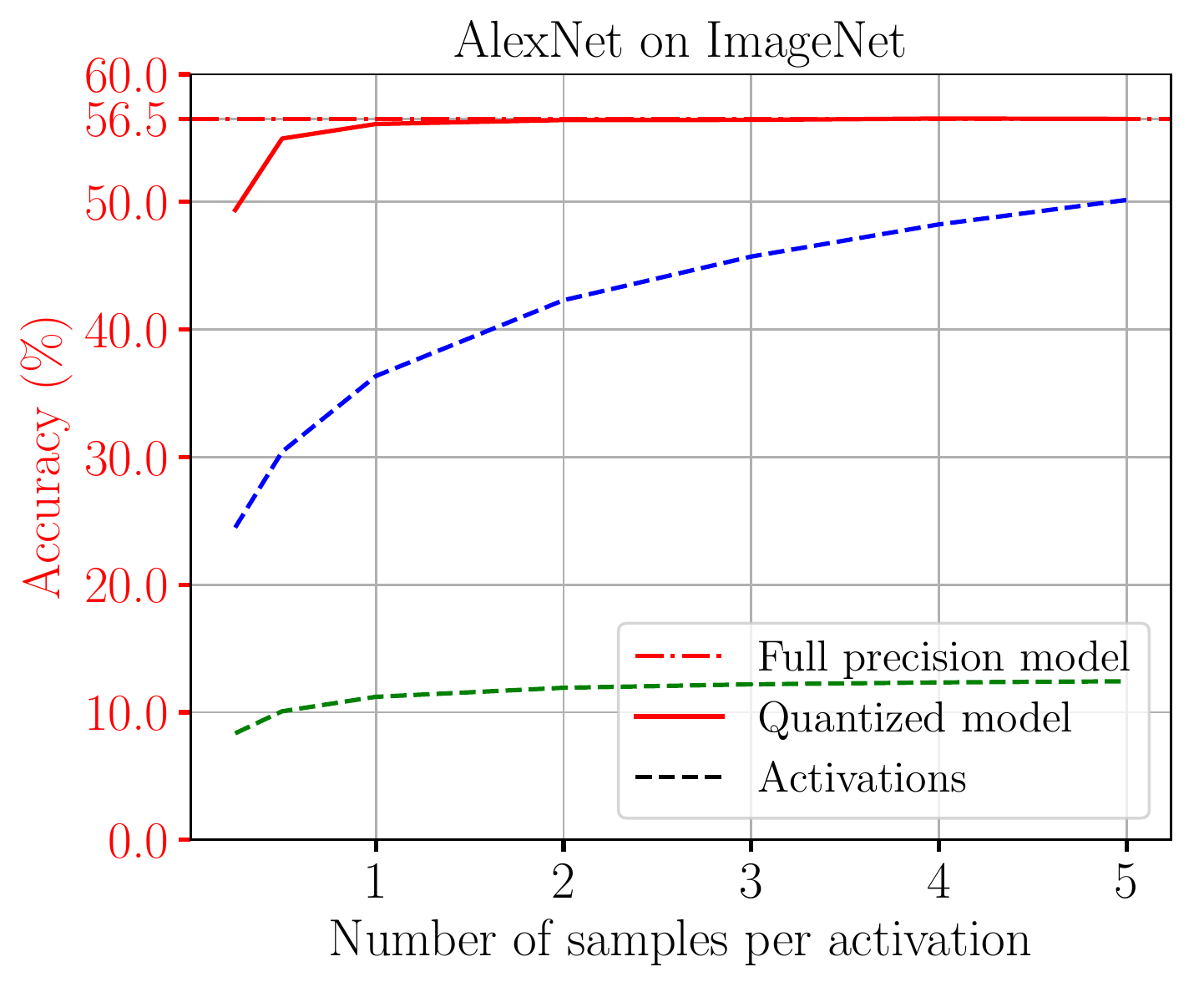}
    \end{subfigure}
    \begin{subfigure}{.305\linewidth}
    	\includegraphics[width=\linewidth]{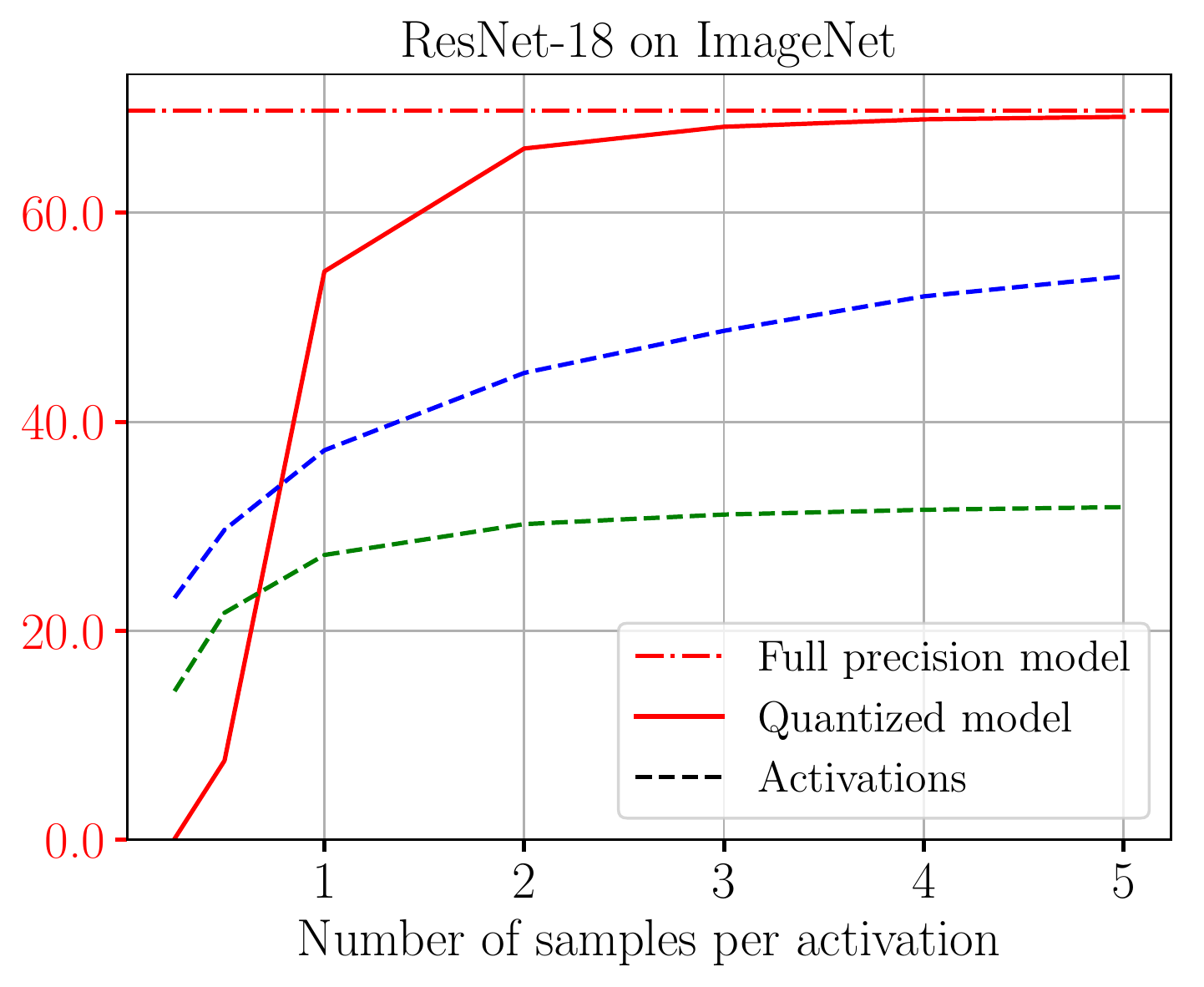}
    \end{subfigure} 
    \begin{subfigure}{.305\linewidth}
    	\includegraphics[width=\linewidth]{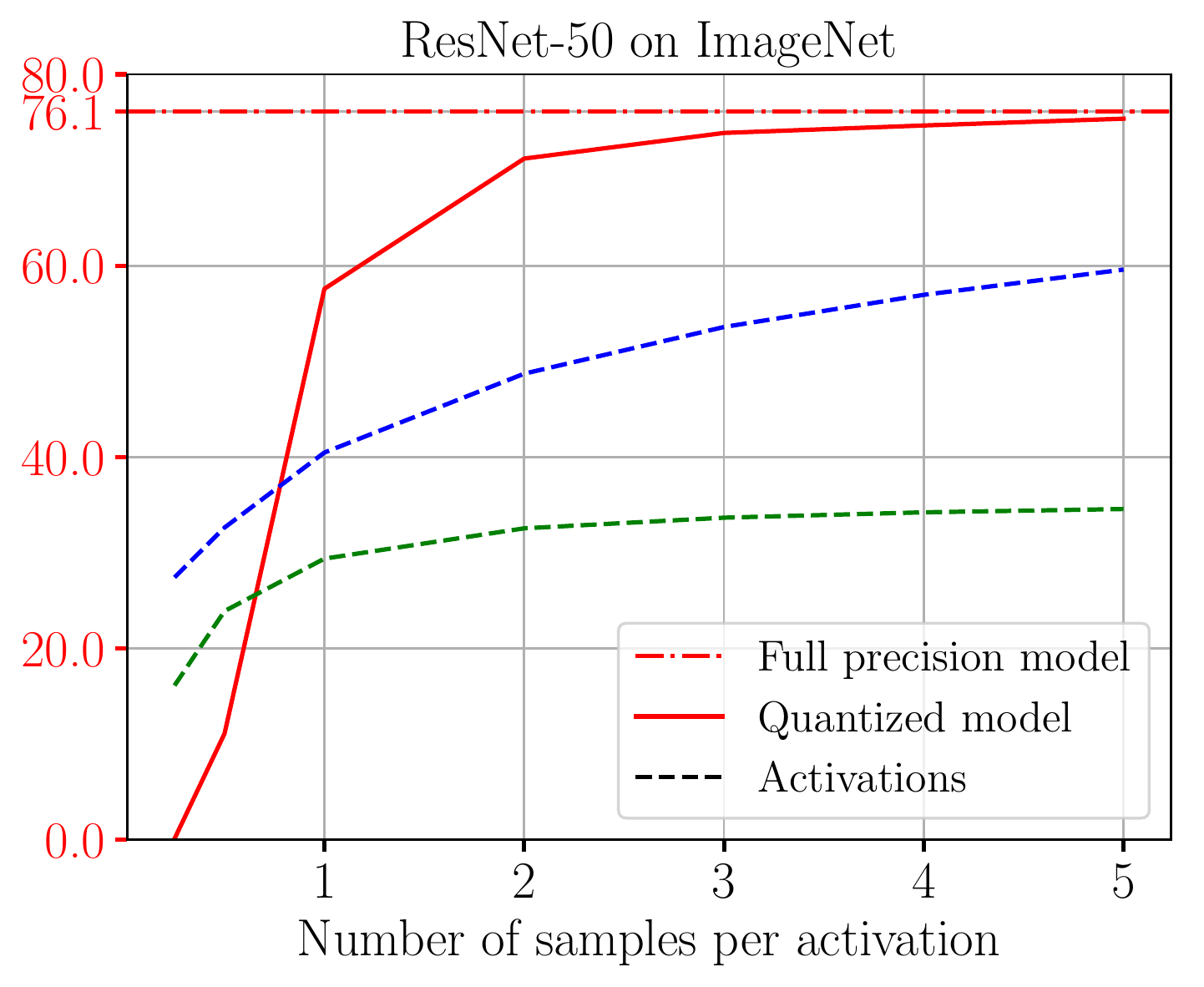}
    \end{subfigure}
    \begin{subfigure}{.062\linewidth}
    	\includegraphics[width=\linewidth]{Images/imagenet/weights/rightax.png}
    \end{subfigure}
    \caption{Quantized activations on ImageNet}
    \label{sup:imagenet_activations}
\end{figure*}

Figures \ref{sup:cifar_activations}, \ref{sup:svhn_activations}, and \ref{sup:imagenet_activations} show the effects of varying the amounts of sampling when quantizing only the activations. We observe less sampling is required to achieve full-precision accuracy when quantizing only the activations when compared to quantizing the weights only.

\section{Effects of Different Sampling Seeds}

\begin{figure*}[ht] 
    \centering
    \begin{subfigure}{.45\linewidth}
        \includegraphics[width=\linewidth]{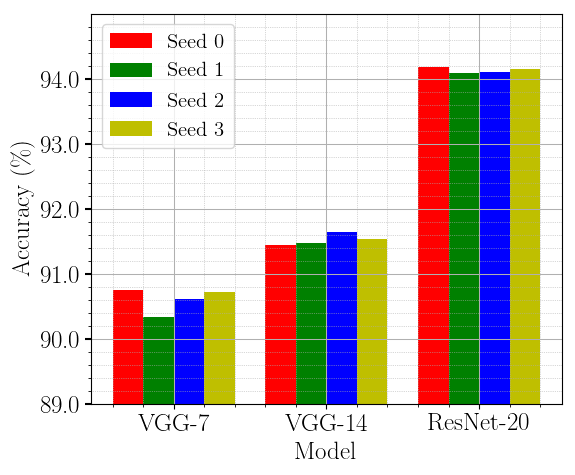} 
        \caption{Quantized weights}
   \end{subfigure} 
    \begin{subfigure}{.45\linewidth}
        \includegraphics[width=\linewidth]{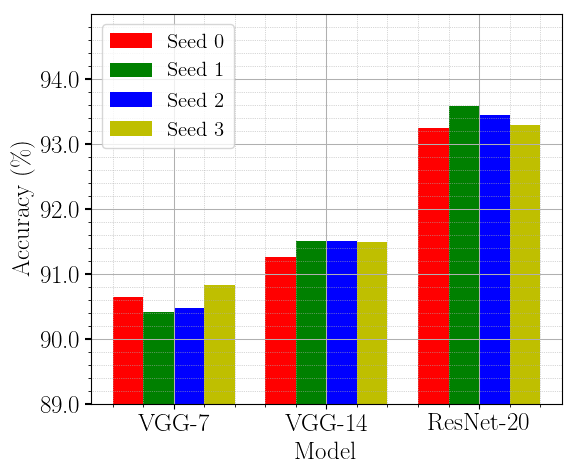}
        \caption{Quantized weights and activations}
    \end{subfigure}
    \caption{Different sampling seeds on CIFAR-10 with $K=1.0$.} 
    \label{fig:different_seeds}
\end{figure*}

In a small experiment on CIFAR-10, we observe that using different sampling seeds can result in up to a $\approx 0.5\%$ absolute variation in accuracy of the different quantized networks (Figure~\ref{fig:different_seeds}). Grid searching over several sampling seeds may then be beneficial to achieve a better quantized model in the end, depending on the use-case.

\end{document}